\documentclass[journal]{IEEEtran}

\usepackage{comment}
\usepackage{fancyvrb}
\usepackage[pdftex]{graphicx}
\usepackage[ragged]{footmisc}
\usepackage{ragged2e}
\usepackage[section]{placeins}
\usepackage{hyperref}
\usepackage{gensymb}

\begin{document}

\title{The SP Theory of Intelligence: Distinctive Features and Advantages}

\author{J Gerard Wolff,~\IEEEmembership{Member,~IEEE}%
\thanks{Dr Gerry Wolff is founder and director of CognitionResearch.org, Menai Bridge, UK. e-mail: jgw@cognitionresearch.org.}%
\thanks{Manuscript received [Month] [day], [year]; revised [Month] [day], [year].}}

\markboth{IEEE Access,~Vol.~00, No.~0, [month]~[year]}%
{J.~G.~Wolff: The SP Theory of Intelligence: Distinctive Features and Advantages}

\maketitle

\begin{abstract}

This paper aims to highlight distinctive features of the {\em SP theory of intelligence}, realised in the SP computer model, and its apparent advantages compared with some AI-related alternatives. Perhaps most importantly, the theory simplifies and integrates observations and concepts in AI-related areas, and has potential to simplify and integrate of structures and processes in computing systems. Unlike most other AI-related theories, the SP theory is itself a theory of computing which can be the basis for new architectures for computers. Fundamental in the theory is information compression via the matching and unification of patterns and, more specifically, via a concept of multiple alignment. The theory promotes transparency in the representation and processing of knowledge, and unsupervised learning of `natural' structures via information compression (DONSVIC). It provides an interpretation of aspects of mathematics and an interpretation of phenomena in human perception and cognition. Abstract concepts in the theory may be realised in terms of neurons and their inter-connections ({\em SP-neural}). These features and advantages of the SP system are discussed in relation to AI-related alternatives: the concept of minimum length encoding and related concepts, how computational and energy efficiency in computing may be achieved, deep learning in neural networks, unified theories of cognition and related research, universal search, Bayesian networks and some other models for AI, IBM's Watson, solving problems associated with big data and in the development of intelligence in autonomous robots, pattern recognition and vision, the learning and processing of natural language, exact and inexact forms of reasoning, representation and processing of diverse forms of knowledge, and software engineering. In conclusion, the SP system can provide a firm foundation for the long-term development of AI and related areas, and, at the same time, it may deliver useful results on relatively short timescales.

\end{abstract}


\section{Introduction}\label{introduction_section}

The {\em SP theory of intelligence} is designed to simplify and integrate observations and concepts across artificial intelligence, mainstream computing, mathematics, and human perception and cognition, with information compression via {\em multiple alignment} as a unifying theme. It is realised in the SP computer model which may be regarded as an early version of the {\em SP machine}. The theory and some of its potential benefits and applications are described in outline in Appendix \ref{outline_of_sp_theory_appendix}, with pointers to where fuller information may be found.

The name `SP' stands for {\em simplicity} and {\em power}, two ideas which are equivalent to information compression  (Appendix \ref{information_compression_appendix}) and thus central in the workings of the SP system, and also in Occam's Razor and the evaluation of scientific theories, as described in Appendix \ref{occams_razor_appendix}.

The SP theory is one of many theories, systems, or schemes that are vying for the attention of researchers in AI and related areas. Since it takes a fair amount of work to fully understand any one of these competing systems, researchers are naturally selective about where they will concentrate their efforts. Accordingly, the main aim of this paper is to present a case for the SP theory.

The beginnings of the argument are simply to show potential benefits and applications of the theory. Much of this has been done already in previous publications referenced in Appendix \ref{outline_of_sp_theory_appendix}.

But this paper goes further. Instead of merely describing what the theory is good for, the paper highlights what is {\em distinctive} about the SP theory compared with alternatives and, more importantly, it argues, immodestly, that in terms of the long-term development of AI-related research, the SP theory has {\em advantages} compared with alternatives, and, at the same time, it has potential in some areas on relatively short timescales. Some of the pros and cons of long- and short-term perspectives in research are discussed briefly in Appendix \ref{long_short_perspectives_appendix}.

In comparing the SP system with others, this paper aims to avoid unnecessary repetition of material that has already been published elsewhere. Trying to include, within this paper, a comprehensive description of the SP system and its applications would not be feasible or reasonable. But to aid understanding and readability, some repetition is unavoidable. Many aspects of the SP system are described in outline or summarised. The paper includes several figures, several of them new, showing output from the SP computer model, with explanations in the text.

The next section provides a broad-brush view of distinctive features of the SP theory, and its strengths. Sections that follow aim to highlight apparent advantages of the SP concepts compared with AI-related alternatives, acknowledging intellectual debts of the SP system, and its shortcomings.

\section{Overview of Distinctive Features and Strengths of the SP Theory}\label{overview_section}

This section is an expanded and revised version of \cite[Section II-G]{sp_autonomous_robots}, summarising distinctive features and strengths of the SP system.

\subsection{Simplification and Integration of Observations and Concepts}\label{ovv_simplification_integration_concepts_section}

As noted in the Introduction and Appendix \ref{outline_of_sp_theory_appendix}, the SP theory aims to simplify and integrate observations and concepts across a broad canvass. Although the theory is not complete (Appendix \ref{future_developments_appendix}), there is now much evidence that the attempt is proving successful---that the SP theory, in accordance with Occam's Razor (Appendix \ref{sp_simplicity_power_evaluation_appendix}), combines relative simplicity with descriptive and explanatory power across a wide range of observations and concepts (Appendix \ref{empirical_conceptual_support_appendix}) and across a wide range of potential benefits and applications (Appendix \ref{benefits_applications_appendix}).

Combining relative simplicity with descriptive and explanatory power in AI-related areas is perhaps the most distinctive feature of the SP theory, and its major strength.

\subsection{Simplification and Integration in Computing Systems}\label{ovv_simplification_integration_computers_section}

The provision of one simple format for knowledge (Appendix \ref{patterns_and_symbols_appendix}), and one framework for the processing of knowledge (Appendix \ref{information_compression_appendix}), promotes an overall simplification of computing systems, including both hardware and software \cite[Section 5]{sp_benefits_apps}.

Those two things also promote seamless integration of diverse kinds of knowledge and diverse aspects of intelligence \cite[Section 7]{sp_benefits_apps}, an integration that appears to be necessary if we are to achieve human-like versatility and adaptability in AI \cite[Section IV-A]{sp_autonomous_robots}.

The SP computer model is not yet a rival to some systems that are dedicated to specific functions. But its long-term potential is likely to be greater than dedicated systems because of its strong theoretical foundations, and because it is designed for the simplification and integration of diverse forms of knowledge, diverse kinds of processing, and diverse aspects of intelligence.

\subsection{The SP Theory Is Itself a Theory of Computing}\label{ovv_sp_as_theory_of_computing_section}

Most other AI-related systems are founded on the concept of `computing' as the workings of the universal Turing machine \cite{turing_1936} or equivalent models such as lamda calculus \cite{church_1941} or Post's canonical system \cite{post_1943}.\footnote{An apparent exception is the concept of a ``neural Turing machine'' \cite{graves_etal_2014}.}

By contrast, {\em the SP theory is itself theory of computing} which appears to be Turing-equivalent as argued in \cite[Chapter 4]{wolff_2006}. The gist of the argument is that the SP machine can do much the same as a Post canonical system, a type of computing system which is recognised as being equivalent to a universal Turing machine.

Another reason for believing that the SP system is likely to have the generality of a universal Turing machine is that the main elements of computer programming can be modelled by the multiple alignment concept, which is itself central in the workings of the SP system (Section \ref{software_engineering_section}).

What is distinctive about the SP theory as a theory of computing is that it provides much of the human-like intelligence that is missing from earlier models (Appendices \ref{empirical_conceptual_support_appendix} and \ref{benefits_applications_appendix}. See also Appendix \ref{sp_simplicity_power_evaluation_appendix}).

\subsection{New Architectures for Computers}\label{new_architectures_section}

The SP theory may provide the foundations for two new architectures for computers:

\begin{itemize}

\item The {\em multiple alignment} framework (Section \ref{ovv_multiple_alignment_section} and Appendix \ref{ma_appendix}) may itself provide the basis for a new kind of computer.

\item {\em SP-neural}---the realisation of the multiple alignment framework in terms of neurons and their inter-connections (Section \ref{ovv_sp-neural_section} and Appendix \ref{sp-neural_appendix}) may provide the basis for a different but related new architecture for computers.

\end{itemize}

The potential significance of these architectures in terms of the computational and energy efficiency of computers is outlined in Section \ref{computational_energy_efficiency_of_computers_section}.

\subsection{Information Compression Via the Matching and Unification of Patterns}\label{ovv_icmup_section}

In trying to cut through some of the mathematical complexities associated with information compression, the SP research programme focuses on a simple, `primitive' idea: that information compression may be understood as a search for patterns that match each other, with the merging or `unification' of patterns that are the same. This will be referred to as `ICMUP' meaning ``information compression via the matching and unification of patterns''. The potential advantage of this approach is that it can help us avoid old tramlines, and open doors to new ways of thinking (\cite[Section 2]{sp_foundations}, Appendix \ref{icmup_appendix}).

\subsection{Information Compression Via Multiple Alignment}\label{ovv_multiple_alignment_section}

More specifically, ICMUP provides the basis for a concept of {\em multiple alignment}, borrowed and adapted from that concept in bioinformatics (Appendix \ref{ma_appendix}). Developing this idea as a framework for the simplification and integration of concepts across a broad canvass has been a major undertaking. Some of the versatility of this concept may be seen in multiple alignments shown in this paper (Figures \ref{class-inclusion_figure}, \ref{part_whole_figure}, \ref{recursion_figure}, \ref{planning_figure}, \ref{occlusion_figure}, \ref{kt_multiple_alignment_figure}, \ref{parsing_figure}, and \ref{learning_alignment_figure}) and in other publications about the SP system.

{\em Multiple alignment is a distinctive and powerful idea in the SP programme of research}. This concept, as it has been developed in the SP programme of research, has the potential to be as significant in computing and cognition as the double helix has proved to be in biological sciences.

Because of the significance of ICMUP and multiple alignment in the SP theory, no attempt has been made in this paper to review or consider the vast literature on information compression via other techniques.

As a general rule, applications in which information compression has a role to play have been developed for specific purposes and do not aspire to simplify and integrate concepts across a broad canvass as the SP system is intended to do.

Section \ref{mle_section} discusses how the SP system relates to minimum length encoding and related topics, and Section \ref{dlnn_information_compression_section} makes some remarks about information compression in deep learning with artificial neural networks.

\subsection{The SP System Is Intrinsically Probabilistic}\label{ovv_sp_is_probabilistic_section}

For reasons outlined in Appendix \ref{ic_prediction_probabilities_appendix}, the SP system is probabilistic at its deepest levels, but the all-or-nothing nature of conventional systems may be imitated if required.

\subsection{Transparency in the Representation and Processing of Knowledge}\label{ovv_transparency_section}

\sloppy By contrast with sub-symbolic approaches to artificial intelligence, and notwithstanding objections to symbolic AI,\footnote{See, for example, ``Hubert Dreyfus's views on artificial intelligence'', {\em Wikipedia}, \href{http://bit.ly/1hGHVm8}{bit.ly/1hGHVm8}, retrieved 2014-08-19.} knowledge in the SP system is transparent and open to inspection, and likewise for the processing of knowledge.

\subsection{Unsupervised Learning and the DONSVIC Principle}\label{ovv_donsvic_section}

A related point is that unsupervised learning in the current version of the SP computer model conforms to the `DONSVIC' principle---{\em The Discovery of Natural Structures Via Information Compression} \cite[Section 5.2]{sp_extended_overview}. Evidence to date suggests that, by contrast with sub-symbolic approaches to artificial intelligence, structures created via unsupervised learning in the SP system will normally be structures that people regard as natural and comprehensible.

\subsection{Mathematics}\label{ovv_mathematics_section}

By contrast with other approaches to artificial intelligence, mainstream computing, or human perception and cognition, the SP theory has quite a lot to say about the nature of mathematics. In brief, it appears that several aspects of mathematics may be understood in terms of ICMUP and, potentially, in terms of multiple alignment (\cite[Chapter 10]{wolff_2006}, \cite{sp_foundations}). Although logic has received less attention in the SP programme of research, it seems likely that similar principles will apply there \cite[Chapter 10]{wolff_2006}.

\subsection{Human Perception and Cognition}\label{ovv_perception_cognition_section}

The SP theory draws extensively on research on human perception and cognition. In particular, it is founded, in part, on research developing computer models of the learning of natural language, underpinned by empirical evidence. This research is summarised in \cite{wolff_1988} and in the web page ``Language learning'' in  \href{http://www.cognitionresearch.org}{www.cognitionresearch.org}, with download links to papers.

\subsection{SP-Neural}\label{ovv_sp-neural_section}

The SP theory includes proposals---{\em SP-neural}---for how abstract concepts in the theory may be realised in terms of neurons and neural processes. The SP-neural proposals (Appendix \ref{sp-neural_appendix}) are significantly different from artificial neural networks as commonly conceived in computer science, and arguably more plausible in terms of neuroscience.

\section{Minimum Length Encoding, Algorithmic Information Theory, and Kolmogorov Complexity}\label{mle_section}

This section and the ones that follow consider how the SP theory relates to a selection of AI-related concepts, emphasising distinctive features of the theory and its apparent advantages compared with alternatives, but also acknowledging shortcomings in the SP system as it is now, and where it has drawn inspiration from earlier work.

As mentioned in Appendix \ref{information_compression_appendix}, information compression in the SP theory may be seen as an example of the principle of {\em minimum length encoding} (MLE) \cite{solomonoff_1964,wallace_boulton_1968,rissanen_1978}. Also, information compression and MLE are closely related to {\em algorithmic information theory} (AIT), and {\em Kolmogorov complexity} (KC) \cite{li_vitanyi_2014}.

Amongst these inter-related areas of study, distinctive features of the SP theory are:

\begin{itemize}

\item Most research on information compression, MLE, AIT, and KC, is founded on the assumption that `computing' is defined by the universal Turing machine. By contrast, the SP theory is itself a theory of computing (Section \ref{ovv_sp_as_theory_of_computing_section}).% NEW: universal Turing machine.

\item By contrast with most research in these areas, there is a central role in the SP theory for ICMUP (Section \ref{ovv_icmup_section}) and, more specifically, the concept of multiple alignment (Section \ref{ovv_multiple_alignment_section}).

\end{itemize}

\section{Computational and Energy Efficiency of Computers}\label{computational_energy_efficiency_of_computers_section}

The book {\em Smart Machines} from IBM \cite{kelly_hamm_2013} argues persuasively that much of the value of big data will be lost because of shortcomings in today's computers,\footnote{In this connection, it has been reported (``Big data: 20 mind-boggling facts everyone must read'', {\em Forbes}, 2015-09-30, \href{http://onforb.es/1YOdT2K}{onforb.es/1YOdT2K}) that only about $0.5\%$ of all data is ever analysed.} and that, while some gains in efficiency may be achieved with optimisations in the design of present-day computers, radically new, brain-like kinds of computer will be needed that are ``many orders of magnitude more energy efficient'' than existing computers:

\begin{quote}

    ``The human brain is a marvel. A mere 20 watts of energy are required to power the 22 billion neurons in a brain that's roughly the size of a grapefruit.\footnote{As noted in \cite{sp_autonomous_robots}, the brain's power consumption is probably about 12.6 watts (``Does thinking really hard burn more calories?'', {\em Scientific American}, 2012-07-18, bit.ly/1qJmCBG), and it may contain as many as 86 billion neurons \cite{herculano-houzel_2012}.} To field a conventional computer with comparable cognitive capacity would require gigawatts of electricity and a machine the size of a football field. So clearly something has to change fundamentally in computing for sensing machines to help us make use of the millions of hours of video, billions of photographs, and countless sensory signals that surround us.~...~Unless we can make computers many orders of magnitude more energy efficient, we're not going to be able to use them extensively as our intelligent assistants.'' \cite[pp.~75, 76 and 88]{kelly_hamm_2013}.

\end{quote}

The two architectures outlined in Section \ref{new_architectures_section} are, together, a fundamentally new approach to the design of computers that draw extensively on research in human cognition and neuroscience, with clear potential to make computers ``many orders of magnitude more energy efficient'' and very much smaller, as described in \cite[Section III]{sp_autonomous_robots} (see also \cite[Section IX]{sp_big_data} and Section \ref{big_data_section}). It appears that there are very few alternatives with that kind of potential.

In brief, the arguments are as follows. Since computation in the SP framework is largely a matter of searching for patterns that match each other, gains in efficiency may be achieved by increasing the efficiency of search. This can be done in four main ways that are not mutually exclusive:

\begin{itemize}

\item Compression of information, which is at the heart of the SP system, reduces the size of the data to be searched.

\item Potentially very large gains in efficiency may be achieved by concentrating searches in areas where success is most likely, exploiting the statistical information that the SP system gathers as a by-product of how it works (Appendix \ref{ic_prediction_probabilities_appendix}).

\item With SP-neural (Appendix \ref{sp-neural_appendix}), there is potential to cut out a lot of searching altogether by making direct connections to structures that are already known (somewhat like making URL links to known web pages in a list of bookmarks or favourites), instead of using search for everything, whether or not it is already known (rather like using an internet search engine to find web pages on all occasions, without taking advantage of shortcuts in a list of bookmarks or favourites).

\item There is also potential to integrate processing and data in the manner of `datacentric' computing \cite[Chapter 5]{kelly_hamm_2013}.

\end{itemize}

The potential of the SP system to make deep cuts in the energy consumption of computers contrasts sharply with the large computational resources used by artificial neural networks with deep learning (Section \ref{dlnn_speed_of_learning_section}).

\section{Deep Learning in Neural Networks}\label{dlnn_main_section}

This section, about deep learning (DL) in artificial neural networks (ANNs), draws extensively on a review by Schmidhuber \cite{schmidhuber_2015}, who has achievements and long experience in the field.

Without in any way wishing to diminish the undoubted successes of DL in ANNs (both of which, together, will be referred to as NNs for short), the aim here is to highlight potential advantages of the SP system. This may seem unduly presumptious since the SP system, unlike some NNs, has not won any competitions and has not been adopted or promoted by any company or incorporated in any products. But for reasons given in the subsections that follow, it appears that the SP system is built on firmer foundations than the current generation of NNs, and its long-term prospects are better.

The great variety of NNs makes it difficult to say things that are true of all of them. For that reason, the subsections that follow attempt to say things that are at least true of the majority.

\subsection{Scope for Adaptation}\label{dlnn_scope_for_adaptation_section}

There is a superficial resemblance between NNs and multiple alignments (especially if the latter are realised as SP-neural, as outlined in Appendix \ref{sp-neural_appendix}) because they both have layers or levels and they both have connections between the levels. But NNs are not multiple alignments and provide much less scope for adaptation:

\begin{itemize}

\item Standardly, the number of layers of an NN and the size of each layer are pre-defined, whereas the number of columns or rows in a multiple alignment, and their sizes, depend entirely on the incoming and stored information from which it is built.\footnote{But in the `Group Method of Data Handling' (GMDH), the number of layers and the number of neurons in each layer depend on the problem being solved \cite[Section 5.3]{schmidhuber_2015}. However, it is evident from a review of research in this area \cite{ivakhnenko_ivakhnenko_1995} that NNs of this type are quite different from multiple alignments in the SP system, and they appear to be much less versatile and adaptable.}

\item Normally, there is just one set of layers in an NN with a structure that is fixed, although its behaviour may be changed via the creation of new links within the structure, or changes in the strengths of links. By contrast, the SP system works by building what is normally a great diversity multiple alignments, each one created by drawing patterns from what is normally quite a small pool of New patterns and what may be a very large pool of Old patterns.

\item Standardly, any given layer in an NN connects only with the layer immediately above (if any) and immediately below (if any).\footnote{An exception would be a fully-recurrent neural network.} By contrast, any given column or row in a multiple alignment may have connections with any other column or row, depending on what the multiple alignment represents.

\item Perhaps most importantly, learning in the SP system (Appendix \ref{ic_unsupervised_learning_appendix}) is quite different from gradualist styles of learning in an NN (Section \ref{dlnn_one-trial_learning_section}). Instead of varying the strengths of links between neurons in a pre-defined structure, the SP system learns by creating Old patterns, which may be derived directly from New patterns or, more commonly, from multiple alignments containing New and Old patterns (\cite[Section 3.9.2 and Chapter 9]{wolff_2006}, \cite[Section 5]{sp_extended_overview}). There is potential for the creation of large numbers of different Old patterns, with a corresponding potential for the learning of diverse kinds of knowledge and skills.

\end{itemize}

\subsection{Biological Validity}\label{dlnn_biological_validity_section}

It is generally recognised that NNs are only vaguely related to biological systems. For example: ``\thinspace`Neural networks are actually very loosely inspired by the brain,' says Oren Etzioni, CEO of the Allen Institute for Artificial Intelligence in Seattle. `They are distributed computing elements, but they're very simple as compared with neurons; the connections are extremely simple as compared with a synapse.'\thinspace''\footnote{``A robot finds its way using artificial `GPS' brain cells'', {\em MIT Technology Review}, 2015-10-19, \href{http://bit.ly/1RVwsxc}{bit.ly/1RVwsxc}.}

Although there are still big gaps in our knowledge about neural structures in the brain, and their functions, it appears that the organisation and workings of the SP system is better supported by available evidence:

\begin{itemize}

\item {\em Models of language learning}. As noted in Section \ref{ovv_perception_cognition_section}, the SP programme of research derives largely from earlier research developing computer models of language learning in children, drawing extensively on relevant empirical evidence.

\item {\em Avoidance of over- or under-fitting}. A related point is that the solution to the problem of over-generalisation and under-generalisation that has been developed in research on language learning is likely to provide a better solution to the problems of over-fitting and under-fitting in NNs than other proposals in that area (Section \ref{dlnn_under_over_generalisation_section}).

\item {\em Cell assemblies}. In SP-neural (Appendix \ref{sp-neural_appendix}, \cite[Chapter 11]{wolff_2006}), abstract concepts in the SP theory map neatly into structures that are quite similar to Hebb's \cite{hebb_1949} concept of a ``cell assembly'', itself derived from neurophysiological evidence.

\item {\em One-trial learning}. By contrast with gradualist styles of learning in NNs, the way in which unsupervised learning is done in the SP system provides an explanation for the phenomenon of one-trial learning and, at the same time, explains why it takes time to learn complex knowledge or skills (Section \ref{dlnn_one-trial_learning_section}).

\item {\em Grandmother cells}. In SP-neural, a concept such as one's grandmother would be represented by a `pattern assembly'---the neural equivalent of an SP pattern. By contrast, the organisation and workings of NNs suggest that the neural representation of a concept such as one's grandmother would normally be distributed across many widely-dispersed neurons and many connections amongst neurons (see also Section \ref{dlnn_distributed_localist_encodings_section}).

    On balance, available evidence suggests that grandmother cells, or something like them, do exist in mammalian brains: 1) it is sometimes suggested that the concept of a grandmother cell or cells is implausible because death of the cell or cells would mean that one could no longer recognise one's grandmother. But that is exactly the kind of thing that can happen with people who have suffered a stroke or are suffering from dementia; and 2) there is relatively direct neurophysiological evidence for the existence of grandmother cells in the brain \cite{gross_2002}.

\end{itemize}

Although the `simple' and `complex' types of neuron discovered by Hubel and Wiesel \cite{hubel_2000} appear to have provided some inspiration for DL concepts \cite[Section 5.2]{schmidhuber_2015}, they may also be seen to provide empirical support for hierarchical structures in the SP system, especially SP-neural (Appendix \ref{sp-neural_appendix}).

\subsection{Learning Paradigms}\label{dlnn_learning_paradigms_section}

Supervised learning, unsupervised learning, and reinforcement learning---three forms of learning with NNs---are, in that connection, normally treated as alternatives with equal status.\footnote{Although Schmidhuber acknowledges that unsupervised learning may facilitate supervised learning and reinforcement learning \cite[Section 4.2]{schmidhuber_2015}}.

By contrast, in the SP perspective, unsupervised learning is seen as a foundation for all other forms of learning, including such things as learning by being told and learning by imitation \cite[Sections V-A.1, V-A.2 and V-J]{sp_autonomous_robots}, and the learning of minor and major skills \cite[Sections V-G to V-I]{sp_autonomous_robots}. The main reasons are, in brief, that:

\begin{itemize}

\item As a matter of ordinary observation, much learning occurs without the benefit of labelled examples, help from a teacher, or carrots and sticks:\footnote{With regard to the last point, it is clear that motivations have an influence on learning---we tend to learn things best if they interest us and if we give them attention. But, contrary to the central dogma of Skinnerian learning theory, it is unlikely that motivations are fundamental in learning.} ``Human and animal learning is largely unsupervised: we discover the structure of the world by observing it, not by being told the name of every object.'' \cite[p.~442]{lecun_etal_2015}.

\item Extraction of redundancy from data, which is central in the SP theory, may be seen to operate not only in unsupervised learning but also in supervised learning---where there is redundancy in the associations between labels and corresponding examples---and in reinforcement learning---where there is redundancy in the associations between actions and corresponding rewards or punishments.

\end{itemize}

Overall, the SP system promises to provide a unifying framework for learning of all kinds, potentially more satisfactory than when different forms of learning are treated separately.

\subsection{Learning From a Single Occurrence or Experience}\label{dlnn_one-trial_learning_section}

Most NNs incorporate some variant of the idea, proposed by Hebb \cite{hebb_1949} and known as `Hebbian' learning, that neurons that repeatedly fire at about the same time will tend to become connected, or for existing connections between them to be gradually strengthened.

This kind of `gradualist' mechanism for learning leads to slow changes in the behaviour of NNs, in keeping with the observation that it normally takes time to learn things like how to talk, or how to play the piano. That correspondence between the workings of NNs and a familiar feature of how we learn may strengthen the belief that NNs are psychologically valid.

But this feature of NNs conflicts with the undoubted fact that we can and often do learn things from a single occurrence or experience. Getting burned once will teach us to be careful with fire. We may retain memories for many years of significant events in our lives that occurred only once. And we may recognise a face that we have seen only briefly, we may recognise music that we have heard only once before, and likewise for films. It is true that we may rehearse things mentally but often it seems that, with little or no rehearsal, we remember things that have been seen or heard only once.

Because the slow strengthening of links between neurons does not account for our ability to remember things after a single exposure, Hebb adopted a `reverberatory' theory for this kind of memory \cite[p.~62]{hebb_1949}. But, as Milner has pointed out \cite{milner_1996}, it is difficult to understand how this kind of mechanism could explain our ability to assimilate a previously-unseen telephone number: for each digit in the number, its pre-established cell assembly may reverberate; but this does not explain memory for the {\em sequence} of digits in the number. We may add that it is unclear how the proposed mechanism would encode a phone number in which one or more of the digits is repeated.

The SP theory provides an explanation, both for learning from a single experience, and for the fact that some kinds of learning are slow:

\begin{itemize}

\item {\em Learning from a single experience}. Learning in the SP system (sketched in Appendices \ref{informal_account_of_sp_system_appendix} and \ref{ic_unsupervised_learning_appendix}) starts by assimilating New information directly, followed by a possible encoding of the information in terms of any existing Old patterns, and the creation of newly-minted Old patterns via information compression \cite[Sections 3.9.2 and 9.2.2]{wolff_2006}, \cite[Section 5.1]{sp_extended_overview}. The taking in of New information, with or without its encoding in terms of existing Old patterns, means that the system can learn from a single exposure to a pattern or event, much like an electronic recording system. This is essentially what is sometimes called `episodic memory'.

\item {\em Slow learning of complex knowledge or skills}. Learning something like a natural language is much more complicated than remembering one's first day at school or when one had a ride on a camel. With the learning of complex knowledge or skills, the main challenge is heuristic search through the vast abstract space of possible knowledge structures to find one or two that are reasonably good (Appendix \ref{heuristic_search_appendix}). The learning of this kind of knowledge---sometimes called `semantic memory'---is necessarily a gradual process.

\end{itemize}

\subsection{Computational Resources, Speed of Learning, and Volumes of Data}\label{dlnn_speed_of_learning_section}

In addition to apparent problems with learning from a single experience, there seem to be related issues with NNs concerning the computational resources and volumes of data they require for learning, and their speed of learning. For example:

\begin{itemize}

\item One news report\footnote{``Google's artificial brain learns to find cat videos'', {\em Wired}, 2012-06-26, \href{http://wrd.cm/18YaV5I}{wrd.cm/18YaV5I}.} describes how an NN with ``16,000 computer processors'' and ``one billion connections'' was exposed to ``10 million randomly selected YouTube video thumbnails'', ``over the course of three days''. Then, ``after being presented with a list of 20,000 different items'', it began to recognize pictures of cats.

\item Another report \cite{edwards_2015} refers to ``...~billions or even hundreds of billions of connections that have to be processed for every image'' and ``Training such a large network requires quadrillions of floating point operations ....''.

\item And ``...~the new millennium brought a DL breakthrough in [the] form of cheap, multiprocessor graphics cards or GPUs.~...~GPUs excel at the fast matrix and vector multiplications required ...~for NN training, where they can speed up learning by a factor of 50 and more.'' \cite[Section 4.5]{schmidhuber_2015}.

\end{itemize}

Although it is true that it takes time for a person to learn his or her native language or languages (Section \ref{dlnn_one-trial_learning_section}) and the human brain contains billions of neurons, the current generation of NNs appears to overlook what can be achieved with ICMUP (Section \ref{ovv_icmup_section} and Appendix \ref{icmup_appendix}), with small amounts of data, and quite modest computational resources:

\begin{itemize}

\item In accordance with the theory developed by Marr and Poggio \cite{marr_poggio_1979} in which ICMUP may be seen to play a central role, a computer program developed by Grimson could discover the hidden image in a random-dot stereogram (\cite{julesz_1971}, \cite[Section 5.1]{sp_vision}) with performance on a late-1970s computer that ``coincides well with that of human subjects'' \cite[Section 5]{grimson_1981}. Although Grimson does not give run times, it looks as if his program finds the hidden image in a random-dot stereogram about as fast as people---normally in less than a minute.

\item With run times of only a few minutes on a PC, the SP computer model, founded on ICMUP, demonstrates unsupervised learning of plausible generative grammars for the syntax of English-like artificial languages (Section \ref{nl_learning_section}). Similar results have been obtained with earlier models of language learning \cite{wolff_1988}, also founded on ICMUP.

\item In a similar vein, Joshua Tenenbaum of MIT has been quoted as saying ``With all the progress in machine learning, it's amazing what you can do with lots of data and faster computers,~...~But when you look at children, it's amazing what they can learn from very little data. ...''\footnote{From ``A learning advance in artificial intelligence rivals human abilities'', {\em New York Times}, 2015-12-10.}

\end{itemize}

It is true that what is being attempted with many NNs is relatively ambitious, but we should not forget that biological neurons are very much slower than electronic components. On balance, there appear to be problems with NNs relating to the computational efficiency of their learning and the volumes of data they require to obtain useful results.

Although researchers with NNs do not normally record the amount of energy required to run them, it is clear that, with the large computational resources and large volumes of data described in the quotes above, the current generation of NNs consume very much more energy than the {\em ca.}~13 watts of the human brain, and the trend appears to be upwards. NNs, as currently conceived, do nothing to solve the problems of energy consumption in computing outlined in Section \ref{computational_energy_efficiency_of_computers_section}. By contrast, there is potential, via information compression and the exploitation of statistical information which the SP system gathers as a by-product of how it works, for very substantial reductions in the amount of energy consumed in AI and, more generally, in computing (Section \ref{computational_energy_efficiency_of_computers_section}, \cite[Section III]{sp_autonomous_robots}).

\subsection{Recognition of Images and Speech}\label{dlnn_recognition_section}

With some qualification (Sections \ref{dlnn_speed_of_learning_section}, \ref{dlnn_easily_fooled_section}, and \ref{dlnn_under_over_generalisation_section}), NNs do well in tasks such as the recognition of images (eg, \cite{yu_etal_2015}) or speech (eg, \cite{dahl_etal_2012}). A particular strength of NNs in this connection is that they may be applied to `raw' digitised data without the need for pre-processing to discover such features as edges or angles \cite[p.~436]{lecun_etal_2015}. But the SP system has potential for the discovery of such features, as discussed in \cite[Section 3]{sp_vision}.

Notwithstanding the strengths of NNs just mentioned, it appears that the SP system provides a firmer foundation for the development of human-like capabilities in computer vision, and perhaps also in the processing of speech. As outlined in Section \ref{pattern_recognition_vision_section}, the SP system has relative strengths and potential in several different aspects of pattern recognition and vision.

\subsection{Deep Neural Networks Are Easily Fooled}\label{dlnn_easily_fooled_section}

A recent report describes how ``We can cause [a deep neural network] to misclassify an image by applying a certain hardly perceptible perturbation.'' \cite[Abstract]{szegedy_etal_2014}. For example, the NN may correctly recognise a picture of a car but may fail to recognise another slightly different picture of a car which, to a person, looks almost identical ({\em ibid.}, Figure 6).

Another report \cite{nguyen_etal_2015} describes how one kind of deep neural network can be fooled quite easily into assigning an image with near certainty to a recognisable class of objects such as `guitar' or `penguin', when people judge the given image to be something like white noise on a TV screen or an abstract pattern containing nothing that resembles a guitar or a penguin or any other object.

Of course, these kinds of failures are a potential problem in any kind of application where recognition needs to be reliable. And without a good theory for how NNs work (Section \ref{dlnn_theoretical_foundations_section}), they may be difficult to weed out.

With regard to the first kind of error---failing to recognise something that is almost identical to what has been recognised---there is already evidence that the SP computer model would not make that kind of mistake. It can recognise words containing errors of omission, commission and substitution (Section \ref{pattern_recognition_vision_section}, \cite[Section 6.2.1]{wolff_2006}), and likewise for diseases in medical diagnosis viewed as pattern recognition \cite[Section 3.6]{wolff_medical_diagnosis} and in the parsing of natural language \cite[Section 4.2.2]{sp_extended_overview}.

No attempt has been made to test experimentally whether or not the SP computer model is prone to the second kind of error---recognising abstract patterns as ordinary objects---but a knowledge of how it works suggests that it would not be.

\subsection{Under-Generalisation and Over-Generalisation}\label{dlnn_under_over_generalisation_section}

An issue with any learning system is its ability to generalise from the data ({\bf I}) that is the basis of its learning, without under-generalisation (`overfitting') or over-generalisation (`underfitting'). If, for example, the system has learned the concept `horse', it should, in its later recognition of horses, not be too closely constrained to recognise only horses that are identical to or very similar to those in {\bf I} (under-generalisation) and, at the same time, it should not make such mistakes as assigning cows, sheep or dogs to the category `horse' (over-generalisation).

\subsubsection{Under-Generalisation in NNs}

It is widely recognised that NNs may suffer from overfitting, and various solutions have been proposed. For example, Srivastava and colleagues \cite{srivastava_etal_2014} suggest that some neurons in an NN, together with their connections, may be randomly dropped from the NN during training, to prevent them co-adapting too much; while Zeng and colleagues \cite{zeng_etal_2013} suggest that, in a multi-stage classifier, unsupervised pre-training and specially-designed stage-wise supervised training can help to avoid overfitting; and Wiesler and colleagues \cite{wiesler_etal_2014} say that they have found that a ``factorized structure'' can be effective against overfitting.

\subsubsection{Over-Generalisation in NNs}

The problem of underfitting in NNs has also drawn attention. For example, Dauphin and Bengio \cite{dauphin_bengio_2013} show how underfitting may arise from the failure of some big neural networks to take full advantage of their computational capacity, and they make suggestions for overcoming the problem; and Ganin and Lempitsky \cite{ganin_lempitsky_2015} describe how a ``two-stage architecture'' can help overcome problems of underfitting.

\subsubsection{Almost Certainly, Information Compression Solves Both Problems}\label{dlnn_generalisation_compression_section}

Without attempting a detailed comparison with alternative accounts of overfitting and underfitting in NNs, the suggestion here is that, in the SP theory, information compression provides a simple, elegant solution for both problems.

In connection with how children learn their first language or languages, information compression can explain how they learn to generalise correctly beyond the language that they have heard---not too little and not too much (\cite[Section 9.5.3]{wolff_2006}, \cite[Section 5.3]{sp_extended_overview}). This includes correction of the over-generalisations (such as ``hitted'' or ``gooses'') that are prominent in the early stages of a child's learning of language. There is evidence that these things can be achieved without the need for correction by a `teacher' or anything equivalent.

In brief, the argument depends on the idea that, when a body of information, {\bf I}, is compressed, the result comprises a {\em grammar}, which we may refer to as {\bf G}, and an {\em encoding} of {\bf I} in terms of {\bf G}, which we may refer to as {\bf E} (Appendix \ref{grammar_encoding_simplicity_power_appendix}). The two things together represent a lossless compression of {\bf I}, without any generalisation. But {\bf G} by itself represents the recurrent or redundant features of {\bf I}, without the non-redundant parts of {\bf I}. As such, it appears to provide for generalisation beyond {\bf I}, without either over-generalisation or under-generalisation.

It seems likely that the same principles would apply to the learning of grammars for the recognition of such things as images or speech, thus solving two problems---the problems of overfitting and underfitting---with one over-arching principle.

The SP system may also help to solve the problem of overfitting in the way that it can recognise patterns via multiple alignment in the face of errors of omission, commission, and substitution (Section \ref{pattern_recognition_vision_section}, \cite[Section 4.2.2]{sp_extended_overview}).

\subsection{Information Compression}\label{dlnn_information_compression_section}

Schmidhuber's review \cite{schmidhuber_2015} contains a short section (4.4) about ``Occam's Razor: compression and minimum description length (MDL)'', and it mentions information compression in some other sections. Although he suggests (in Section 5.10) that ``much of machine learning is essentially about compression'', the overall thrust of the review is that information compression is merely one of several ``recurring themes'' in deep learning, without any great significance.

By contrast, information compression is fundamental in the SP theory, running through it like {\em Blackpool} in a stick of rock, in its foundations (\cite{sp_foundations}, Appendix \ref{sp_foundations_and_scope_appendix}), in the matching and unification of patterns (Appendix \ref{icmup_appendix}), in the building of multiple alignments (Appendix \ref{ma_appendix}), and in unsupervised learning (Appendix \ref{ic_unsupervised_learning_appendix}).

In view of evidence for the importance of information compression in intelligence, computing, and mathematics (Appendix \ref{sp_foundations_and_scope_appendix}), it appears that the peripheral status of information compression in the design and operation of NNs weakens them conceptually in comparison with the SP system.

\subsection{Transparency in the Representation and Processing of Knowledge}\label{dlnn_transparency_section}

A problem with NNs is that there is considerably uncertainty about how they represent knowledge and how they process it:

\begin{quote}

    ``...~we actually understand surprisingly little of why certain models work and others don't.~...~One of the challenges of neural networks is understanding what exactly goes on at each layer.'' \cite{mordvintsev_etal_2015}.

    ``...~no one knows how neural networks come up with their answers.~...~A programmer need adjust only the number of nodes and layers to optimise how it captures relevant features in the data. However, since it's impossible to tell exactly how a neural network does what it does, this tweaking is a matter of trial and error.''\footnote{``The rapid rise of neural networks and why they'll rule our world'', {\em New Scientist}, 2015-07-08, \href{http://bit.ly/1IkbbuC}{bit.ly/1IkbbuC}.}

\end{quote}

With regard to the first quote, it is true that, as described in the blog, NNs can be made to reveal some of their knowledge. But, while many of the resulting images have artistic appeal, they are not transparent representations of knowledge (see also Section \ref{dlnn_class-inclusion_part-whole_section}), it's not clear how they are learned or how they function in such tasks as recognition, and they certainly do not conform to the DONSVIC principle (Section \ref{ovv_donsvic_section}).

By contrast with these uncertainties:

\begin{itemize}

\item In the SP system, all kinds of knowledge, including those detailed in Section \ref{representation_processing_knowledge_section}, are represented transparently as SP patterns.

\item In the SP computer model, an audit trail can be provided for all processing, including the building of multiple alignments and the creation of grammars.

\item In the SP system, it is anticipated that unsupervised learning will conform to the DONSVIC principle (Section \ref{ovv_donsvic_section}, \cite[Section 5.2]{sp_extended_overview}), and this is confirmed by evidence from the SP computer model. To the extent that this remains true in future versions of the model, structures created via unsupervised learning in the SP model are likely to be transparent and comprehensible by people.

\end{itemize}

\subsubsection{Distributed or Localist Encodings}\label{dlnn_distributed_localist_encodings_section}

There is further uncertainty about whether knowledge in an NN is represented and processed in a `distributed' or `localist' scheme. The dominant view is that, in neural networks, knowledge of a concept such as one's grandmother is encoded in neurons that are widely-distributed, with links between them (Section \ref{dlnn_biological_validity_section}). In this `sub-symbolic' view, it would not be possible to identify any single neuron or local cluster of neurons that represent any given concept: ``there is no distinction between individual high level units and random linear combinations of high level units, according to various methods of unit analysis'', suggesting that ``it is the entire space of activations, rather than the individual units, that contains the bulk of the semantic information'' \cite[Abstract]{szegedy_etal_2014}.

But some researchers suggest that ``it is possible to train [artificial] neurons to be selective for high-level concepts~...~In our experiments, we obtained neurons that function as detectors for faces, human bodies, and cat faces by training on random frames of YouTube videos.'' \cite[Conclusion]{le_etal_2013}. This research appears to support the `localist' or `symbolic' view that a concept such as one's `grandmother' may be represented by a single neuron or, perhaps, a small cluster of neurons ({\em cf.} Section \ref{dlnn_biological_validity_section}).

Although the SP theory may turn out to be wrong empirically, it is at least clear on this issue. As noted in Section \ref{dlnn_biological_validity_section}, it is very much in the localist camp. In SP-neural, it is envisaged that knowledge of concepts in the form of SP patterns is recorded on the cortex very much like writing on a sheet of paper (Appendix \ref{sp-neural_appendix}). As with writing on a sheet of paper, any one section or paragraph, in one location, may be seen to describe a concept, although it may contain pointers to supporting information elsewhere.

\subsubsection{Class-Inclusion Hierarchies and Part-Whole Hierarchies}\label{dlnn_class-inclusion_part-whole_section}

With regard to hierarchical structures, there is further uncertainty about whether NNs discover:

\begin{itemize}

\item Class-inclusion hierarchies:

\begin{quote}

    ``Deep-learning methods are representation-learning methods with multiple levels of representation, obtained by composing simple but non-linear modules that each transform the representation at one level (starting with the raw input) into a representation at a higher, slightly more abstract level.'' \cite[p.~436]{lecun_etal_2015}.

    ``Hidden layers: these learn more abstract representations as you head up.''\footnote{Slide 3 in ``Deep learning for NLP (without magic)'', slide show by R.~Socher and C.~Manning, dated 2013, retrieved 2015-07-25, \href{http://stanford.io/1bmBsKK}{stanford.io/1bmBsKK}.}).

\end{quote}

\item Or part-whole hierarchies:

\begin{quote}

    ``An image ...~comes in the form of an array of pixel values, and the learned features in the first layer of representation typically represent the presence or absence of edges at particular orientations and locations in the image. The second layer typically detects motifs by spotting particular arrangements of edges, regardless of small variations in the edge positions. The third layer may assemble motifs into larger combinations that correspond to parts of familiar objects, and subsequent layers would detect objects as combinations of these parts.'' \cite[p.~436]{lecun_etal_2015}.

    ``...~the first layer maybe looks for edges or corners. Intermediate layers interpret the basic features to look for overall shapes or components, like a door or a leaf. The final few layers assemble those into complete interpretations ...~such as entire buildings or trees.'' \cite{mordvintsev_etal_2015}.

\end{quote}

\end{itemize}

\noindent and, either way, the representation is obscure.

While it is clear that a human face is part of a human body (Section \ref{dlnn_distributed_localist_encodings_section}), it seems reasonable to assume that, in either a distributed or localist scheme (Section \ref{dlnn_distributed_localist_encodings_section}), a concept such as `horse', `cow', or `sheep', would be most fully encoded in the highest layer of an NN. If that is accepted, the question arises how the NN would encode knowledge of something like an agricultural exhibition which is likely to contain representations of all three of the concepts mentioned, and with multiple instances of each one of them. Likewise, we may ask how such concepts may be encoded as examples of abstractions such as `mammal' or `vertebrate'. It is not clear how concepts that are most fully encoded in the top layer of an NN could ever be part of something that is larger or more abstract.

\begin{figure*}[!htbp]
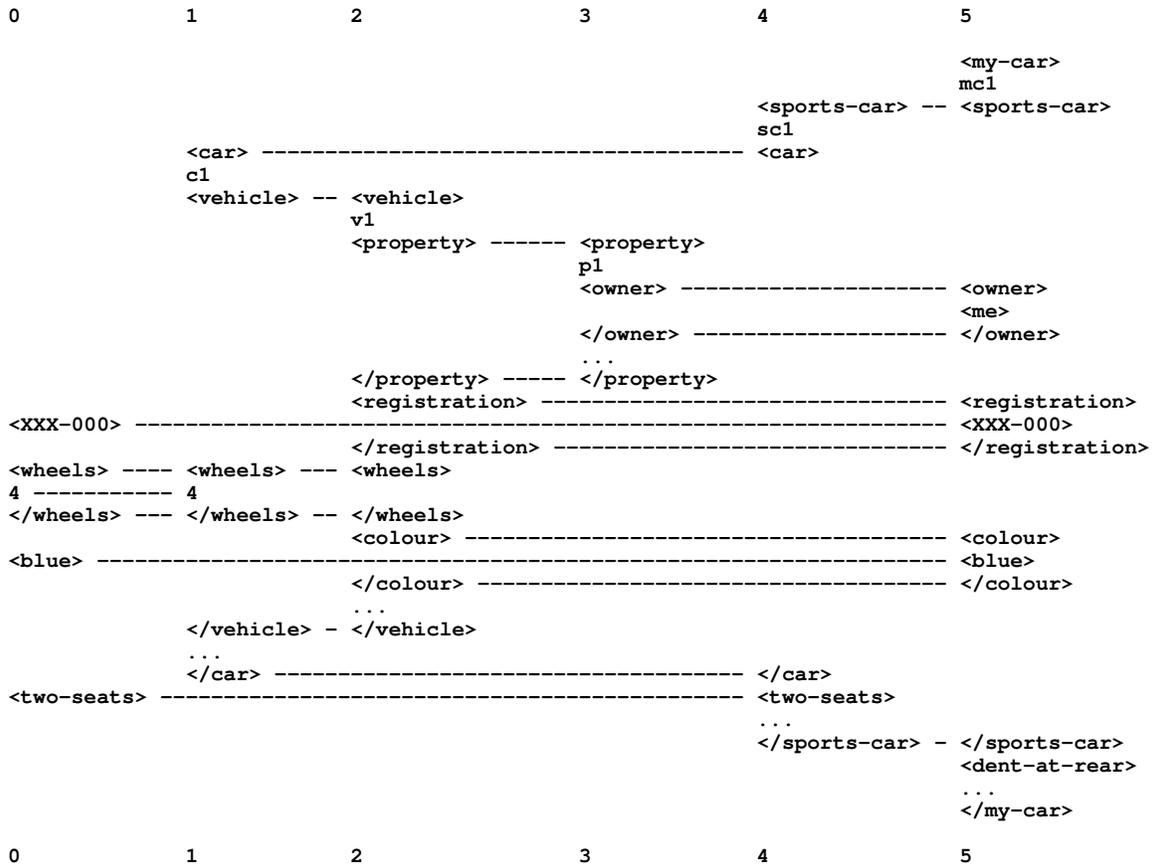

\fontsize{08.00pt}{08.60pt}
\centering
{\bf
\begin{BVerbatim}
0             1            2                 3             4               5

                                                                           <my-car>
                                                                           mc1
                                                           <sports-car> -- <sports-car>
                                                           sc1
              <car> -------------------------------------- <car>
              c1
              <vehicle> -- <vehicle>
                           v1
                           <property> ------ <property>
                                             p1
                                             <owner> --------------------- <owner>
                                                                           <me>
                                             </owner> -------------------- </owner>
                                             ...
                           </property> ----- </property>
                           <registration> -------------------------------- <registration>
<XXX-000> ---------------------------------------------------------------- <XXX-000>
                           </registration> ------------------------------- </registration>
<wheels> ---- <wheels> --- <wheels>
4 ----------- 4
</wheels> --- </wheels> -- </wheels>
                           <colour> -------------------------------------- <colour>
<blue> ------------------------------------------------------------------- <blue>
                           </colour> ------------------------------------- </colour>
                           ...
              </vehicle> - </vehicle>
              ...
              </car> ------------------------------------- </car>
<two-seats> ---------------------------------------------- <two-seats>
                                                           ...
                                                           </sports-car> - </sports-car>
                                                                           <dent-at-rear>
                                                                           ...
                                                                           </my-car>

0             1            2                 3             4               5
\end{BVerbatim}
}
\caption{A multiple alignment illustrating recognition at multiple levels of abstraction, as described in the text.}
\label{class-inclusion_figure}
\end{figure*}

\begin{figure*}[!htbp]
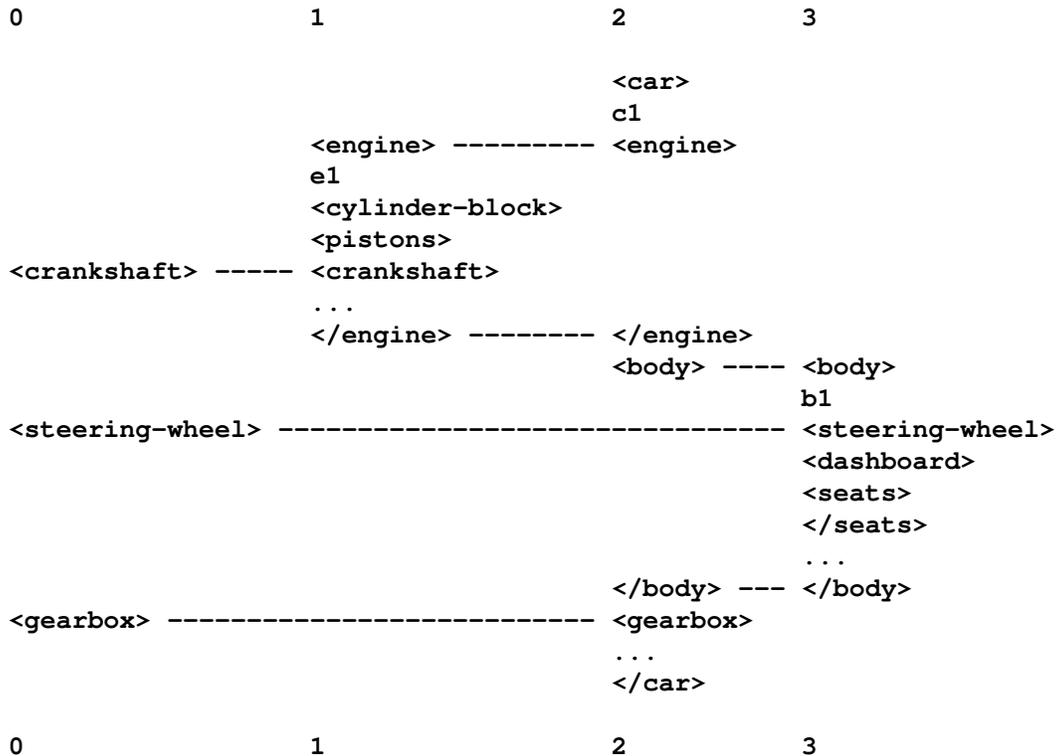

\fontsize{10.00pt}{12.00pt}
\centering
{\bf
\begin{BVerbatim}
0                  1                  2           3

                                      <car>
                                      c1
                   <engine> --------- <engine>
                   e1
                   <cylinder-block>
                   <pistons>
<crankshaft> ----- <crankshaft>
                   ...
                   </engine> -------- </engine>
                                      <body> ---- <body>
                                                  b1
<steering-wheel> -------------------------------- <steering-wheel>
                                                  <dashboard>
                                                  <seats>
                                                  </seats>
                                                  ...
                                      </body> --- </body>
<gearbox> --------------------------- <gearbox>
                                      ...
                                      </car>

0                  1                  2           3
\end{BVerbatim}
}
\caption{A multiple alignment illustrating recognition of a car via its parts and sub-parts, as described in the text.}
\label{part_whole_figure}
\end{figure*}

By contrast with these uncertainties with NNs about the representation of class-inclusion hierarchies or part-whole hierarchies, the SP system can represent both types of hierarchy, with a clear distinction between the two. This can be seen in the class hierarchies in Figure \ref{class-inclusion_figure} and in \cite[Figure 6.7]{wolff_2006}, and in the part-whole hierarchies in Figure \ref{part_whole_figure} and in \cite[Figure 6.8]{wolff_2006}). At the same time it provides for seamless integration of both kinds of hierarchy, as illustrated in \cite[Figure 16]{sp_extended_overview}. Any concept, of any size or level of abstraction, may always be embedded in something larger or more abstract.

The SP system also provides the means of representing class-inclusion heterarchies (cross-classification) and part-whole heterarchies. It appears that this is well beyond anything that can be done with NNs.

Figure \ref{class-inclusion_figure} shows how the SP computer model may recognise a car at multiple levels of abstraction. The figure shows the best multiple alignment created by the SP computer model---the one yielding most compression---with a New pattern (column 0) and a set of Old patterns representing categories related to cars (columns 1 to 5). Column 5 shows that the unknown entity is ``my car'', and other columns show that it is a sports car (column 4), a car (column 1), and a vehicle (column 2). The order of the categories across the columns has no special significance---it depends purely on how the multiple alignment was built up.

Figure \ref{part_whole_figure} shows how, in a similar way, the SP system may, via a multiple alignment, model the recognition of an unknown entity in terms of its parts and sub-parts. In this example, the features of the unknown entity are shown in column 0. That the entity is a car is shown in column 2, with a match for one of its features---a gearbox. Column 1 shows one part of a car---the engine---with a match for one of its parts---the crankshaft. And column 3 represents the body of the car with a match for one of its parts---the steering wheel.

\subsubsection{Iteration and Recursion}\label{dlnn_iteration_recursion_section}

Related to issues just discussed are questions about how NNs may encode repeated instances of this or that category (such as the many instances of horses, cows, and sheep that one would expect to see at an agricultural exhibition (Section \ref{dlnn_class-inclusion_part-whole_section})), and how NNs may encode the kinds of recursive structures that are prominent in natural language, such as {\em This is the horse and the hound and the horn, That belonged to the farmer sowing his corn, That kept the rooster that crowed in the morn, ...}.

At first sight, one might think that `recurrent neural networks' would be an appropriate means of modelling recursion in natural language. But, as noted in Section \ref{dlnn_nl_parsing_production_section}, these kinds of NNs do not, apparently, respect the word, phrase, clause and sentence structures in natural language that have been prominent in generative and computational linguistics for many years.

By contrast with uncertainties about how NNs may handle recursive structures, the SP theory provides crisp answers:

\begin{itemize}

\item Any given SP pattern may appear one, two, or more times within any one multiple alignment. One example is how the pattern representing a noun phrase (`\texttt{NP}') appears twice in the multiple alignment shown in Figure \ref{parsing_figure}: once in column 2 and again in column 11. Another example is shown in Figure \ref{recursion_figure}, below, where the pattern `\texttt{ri}' appears in columns 5, 7, and 9.

    In the interpretation of any given body of data, there should be no difficulty arising from the occurrence of multiple instances of any given category. There is more detail in \cite[Section 3.4.6]{wolff_2006} about how a given pattern may appear more than once in any multiple alignment.

\item Multiple alignments can accommodate recursive structures, as illustrated in Figures \ref{recursion_figure} and \ref{kt_multiple_alignment_figure}.

\item Recursion may also be accommodated in SP-neural, as illustrated in \cite[Figure 11.10]{wolff_2006}.

\end{itemize}

With regard to Figure \ref{recursion_figure}, native speakers of English know that, in a phrase like {\em the very ...~expensive jewellery}, the word {\em very} may, in principle, be repeated any number of times---although it would be considered poor style to use it more than 4 or 5 times. In the figure, this kind of recursion may be seen in repeated appearances of the self-referential pattern `\texttt{ri ri1 i \#i ri \#ri \#ri}'. It is self-referential because the symbols `\texttt{ri ...~\#ri}' at the beginning and end of the pattern can be aligned with the symbols `\texttt{ri \#ri}' in the body of another appearance of the pattern.

\begin{figure*}[!htbp]
\fontsize{08.00pt}{09.60pt}
\centering
{\bf
\begin{BVerbatim}
0           1           2     3           4      5     6      7     8      9     10

                        np
                        d ------------------------------------------------------ d
                                                                                 d1
the ---------------------------------------------------------------------------- the
                        #d ----------------------------------------------------- #d
                        ri --------------------- ri
                                                 ri1
                                          i ---- i
                                          i1
very ------------------------------------ very
                                          #i --- #i
                                                 ri --------- ri
                                                              ri1
                                                       i ---- i
                                                       i1
very ------------------------------------------------- very
                                                       #i --- #i
                                                              ri --------- ri
                                                                           ri1
                                                                    i ---- i
                                                                    i1
very -------------------------------------------------------------- very
                                                                    #i --- #i
                                                                           ri
                                                                           #ri
                                                              #ri -------- #ri
                                                 #ri -------- #ri
                        #ri -------------------- #ri
                        a --- a
                              a1
expensive ------------------- expensive
                        #a -- #a
            n --------- n
            n1
jewellery - jewellery
            #n -------- #n
                        #np

0           1           2     3           4      5     6      7     8      9     10
\end{BVerbatim}
}
\caption{A multiple alignment from the SP computer model showing recursion, as described in the text.}
\label{recursion_figure} % (Section \ref{})
\end{figure*}

\subsection{The Processing of Natural Language}\label{dlnn_nl_processing_section}

A recent review of research in deep learning \cite{lecun_etal_2015} suggests that ``deep learning has produced extremely promising results for various tasks in natural language understanding, particularly topic classification, sentiment analysis, question answering and language translation.'' (p.~436). This section first examines some research in the application of NNs to natural language, and then, in Section \ref{dlnn_nl_discussion_section}, assesses their prospects in this area in comparison with the strengths and potential of the SP system.

To anticipate a little, it seems that NNs may indeed produce some useful results in the processing of natural language but, taking a longer view, the SP system promises more precision and higher standards of performance. As one commentator has remarked: ``Deep learning's usefulness for speech recognition and image detection is beyond doubt, but it's still just a guess that it will master language and transform our lives more radically.'' \cite[p.~73]{simonite_2015}. It seems unlikely that NNs will, for example, ever be able to represent and process the subtle structure of inter-locking dependencies in English auxiliary verbs, as the SP system can do (\cite[Section 5.5]{wolff_2006}, \cite[Sections 8.2 and 8.3]{sp_extended_overview}).

\subsubsection{Parsing and Production of Natural Language}\label{dlnn_nl_parsing_production_section}

In addressing issues in the parsing of natural language by NNs, Socher and colleagues \cite{socher_etal_2013} employ the concept of `Compositional Vector Grammar', but their results are not as clear or straightforward as one might wish, something which two of them (Socher and Manning) acknowledge in a slide show:

\begin{quote}

    ``{\em Concern: problem with model interpretability}. No discrete categories or words, everything is a continuous vector. We'd like have symbolic features like NP, VP, etc. and see why their combination makes sense.''\footnote{Slide 201 in ``Deep learning for NLP (without magic)'', slide show by R.~Socher and C.~Manning, dated 2013, retrieved 2015-07-27, \href{http://stanford.io/1bmBsKK}{stanford.io/1bmBsKK}.}

\end{quote}

\noindent Notwithstanding some qualifying remarks, their concern remains valid.

By contrast, the SP system has several strengths in both the analysis and production of natural language (Section \ref{nlp_section}). In particular, the system respects and fully represents the kinds of hierarchical structure that have long been recognised in generative linguistics, which are quite different from the kinds of cyclic representations of language sequences by recurrent neural networks described in, for example, \cite{vinyals_etal_2014,sutskever_etal_2014}.

\subsubsection{Translation Between Languages}\label{dlnn_nl_translation_section}

In some recent work (eg, \cite{jean_etal_2015,sutskever_etal_2014,schwenk_2012}) NNs have been applied to the already-established technique of translating between languages by mapping phrases in the source language to corresponding phrases in the target language.

The main problem here is that, while phrase-to-phrase or syntax-to-syntax systems can produce useful translations, and may generalise usefully from the training data \cite{schwenk_2012}, it is unlikely that they will ever attain the standard of a good human translator. To reach that `Holy Grail', it appears to be essential to translate the source language into a representation of its meaning, and then translate this `interlingua' into the target language. This appears to be what people do, it is in any case needed for high-quality `understanding' of natural language (Sections \ref{dlnn_nl_understanding_section} and \ref{nlp_section}), and it can reduce substantially the number of mappings that are are required.

In the quest for that higher standard of translation, it appears that the SP system has more potential than NNs: it has clear potential for the learning of the syntax of natural language, for the learning of semantic structures, and for learning syntactic-semantic structures (Section \ref{nl_learning_section}).

\subsubsection{Understanding of Natural Language}\label{dlnn_nl_understanding_section}

Zhang and LeCun \cite[Section 1]{zhang_lecun_2015} write: ``In this article we show that text understanding can be handled by a deep learning system without artificially embedding knowledge about words, phrases, sentences or any other syntactic or semantic structures associated with a language.'' But this claim needs to be qualified:

\begin{itemize}

\item In the examples that are given, what is learned is essentially associations between short pieces of text such as questions and answers in the {\em Yahoo!~Answers} website. This notion of `understanding' falls a long way short of, for example, the understanding that a motor engineer would have for an expression like ``gear box'', or that a musician would have for an expression like ``Handel's Messiah''. For the great majority of `content' words or phrases in any language, the full meaning is a rich structure of knowledge, and, normally, most of it is not about words or combination of words.

\item With the very simplistic concept of `understanding' that is employed in \cite{zhang_lecun_2015}, it may indeed be possible to find associations between words and `meanings' without worrying about the syntactic and semantic structure of language. But for a thorough understanding of, for example, a legal, philosophical or scientific argument, it seems unlikely that we can by-pass the need for a good knowledge of such structures.

\end{itemize}

What is perhaps a better example is described in \cite{vinyals_etal_2014}, where NN techniques for syntax-to-syntax translation are borrowed and converted to an image-to-syntax scheme, where ``image'' means the output of an NN that recognises images and ``syntax'' means a natural language caption for each image. In this case, it is perhaps more reasonable to describe each image as the `meaning' of the corresponding caption. But we would still be a long way from the richly-structured forms of knowledge which provide most of the meanings for natural language.

In summary, it seems that, compared with NNs, the SP system provides a much sounder basis for learning those rich, complex structures of knowledge which are the meanings of natural language and which need to be in place if we are to attain anything like human-level `understanding' of natural language. It seems unlikely that NNs will ever be able to meet the ``The Winograd Schema Challenge'' \cite{levesque_etal_2012}.\footnote{Winograd's original example \cite[p.~33]{winograd_1972} demonstrates the subtlety of natural language with two sentences: {\em The city councilmen refused the demonstrators a permit because they feared violence} and {\em The city councilmen refused the demonstrators a permit because they advocated revolution}. In the first case, we understand that the word {\em they} refers to the city councilmen, while in the second case it seems more reasonable to assume that {\em they} is a reference to the demonstrators. Our understandings in the two cases depend on extensive knowledge of how things happen in the world and the psychology of people, including city councilmen and demonstrators.}

\subsubsection{Unsupervised Learning of Natural Language}\label{dlnn_nl_learning_section}

Zhang and LeCun \cite[Section 1]{zhang_lecun_2015} also write: ``...~we hypothesize that when trained from raw characters, temporal ConvNet is able to learn the hierarchical representations of words, phrases and sentences in order to understand text.'' (Section 1). In view of the problems of transparency with NNs (Section \ref{dlnn_transparency_section}), uncertainties about distributed or localist encodings (Section \ref{dlnn_distributed_localist_encodings_section}), uncertainties about class-inclusion hierarchies and part-whole hierarchies (Section \ref{dlnn_class-inclusion_part-whole_section}), and the failure of NNs to respect the DONSVIC principle (Section \ref{dlnn_transparency_section}, introduction), this hypothesis appears to be over-optimistic.

By contrast, the SP system has already demonstrated unsupervised learning of segmental structures, classes of structure, and abstract patterns, and it has good prospects for further development (Section \ref{nl_learning_section}).

\subsubsection{Discussion}\label{dlnn_nl_discussion_section}

There is little doubt that, in natural language processing, NNs will yield some useful new applications on short to medium timescales. But, because of the several problems with NNs, described elsewhere in Section \ref{dlnn_main_section}, it is unlikely that they can match human capabilities with natural language. In particular, they are unlikely to make much impact on the difficult problems that need to be solved to reach that standard: learning syntactic structures, learning the kinds of knowledge structures that provide the meanings for language, and learning syntactic-semantic connections.

It appears that, taking a long view, the SP system provides a much sounder basis for the understanding, production, translation, and learning of natural language, with firmer theoretical foundations.

\subsection{Reasoning, Programming, and Other Kinds of `Symbolic' Processing}\label{dlnn_symbolic_ai_section}

\begin{quote}

    ``As regards knowledge representation in the brain, one of the key challenges is to understand how neural activations, which are widely distributed and sub-symbolic, give rise to behavior that is symbolic, such as language and logical reasoning.'' \cite[Section 2]{davilagarcez_etal_2015}.

    ``[Deep learning techniques] have no obvious ways of performing logical inferences ...''\footnote{``Is `deep learning' a revolution in artificial intelligence?'', {\em The New Yorker}, Gary Marcus, 2012-11-25, \href{http://nyr.kr/1Be7S22}{nyr.kr/1Be7S22}.}

\end{quote}

This section briefly reviews aspects of AI and other areas of computing which seem to pose problems for NNs, and where the SP system is relatively strong. In broad terms, these seem to be areas where the `symbolic' tradition has proved to be relatively successful.

Some years ago, there was quite a lively interest in how NNs might do reasoning (eg, \cite{levine_aparicio_1994}) but that interest seems to have subsided, probably because NNs really are not well suited to much more than the relatively simple kinds of inference that, in pattern recognition, correct errors of omission, commission, or substitution.

By contrast, the SP system demonstrates several kinds of reasoning, with clear potential for further development (Section \ref{exact_inexact_reasoning_section}).

\begin{figure}[!htbp]
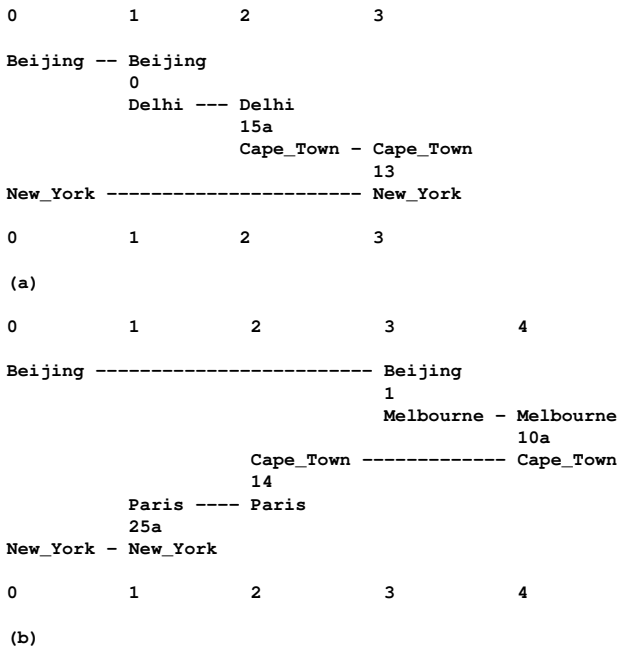

\fontsize{07.00pt}{08.40pt}
\centering
{\bf
\begin{BVerbatim}
0          1         2           3

Beijing -- Beijing
           0
           Delhi --- Delhi
                     15a
                     Cape_Town - Cape_Town
                                 13
New_York ----------------------- New_York

0          1         2           3

(a)

0          1          2           3           4

Beijing ------------------------- Beijing
                                  1
                                  Melbourne - Melbourne
                                              10a
                      Cape_Town ------------- Cape_Town
                      14
           Paris ---- Paris
           25a
New_York - New_York

0          1          2           3           4

(b)
\end{BVerbatim}
}
\caption{Two multiple alignments produced by the SP computer model showing how, with appropriate data, the system may plot alternative routes between two cities, as described in the text. This figure has been adapted, with permission, from Figures 8.1 and 8.5 in \cite{wolff_2006}.}
\label{planning_figure}
\end{figure}

Similar things may be said about areas of AI which appear to be problematic for NNs and where symbolic AI and the SP system are relatively strong, such as planning (see Figure \ref{planning_figure}, below, and \cite[Chapter 8]{wolff_2006}), problem solving (\cite[Chapter 8]{wolff_2006})), and unsupervised learning of language (Section \ref{nl_learning_section}, \cite[Chapter 9]{wolff_2006}, \cite[Section 5]{sp_extended_overview}).

Again, it is difficult to envisage how NNs might be used for the kind of `programming' which is the mainstay of software engineering as it is practiced now. By contrast, the SP system has clear potential for application in that area (Section \ref{software_engineering_section}).

There seem to be three main reasons for the relative strength of the SP system in the areas mentioned:

\begin{itemize}

\item Unlike NNs, there is transparency in both the representation and processing of knowledge (Sections \ref{ovv_transparency_section} and \ref{dlnn_transparency_section}).

\item Within the multiple alignment framework, it is possible to model concepts from mainstream computing and the symbolic tradition of AI such as {\em variable}, {\em value}, and {\em type} (Section \ref{exact_inexact_reasoning_section}).

\item The multiple alignment framework is much more adaptable than the deep learning framework (Section \ref{dlnn_scope_for_adaptation_section}).

\end{itemize}

As an example of the kind of thing that the SP system can do, Figure \ref{planning_figure} shows how multiple alignments produced by the SP computer model may serve to plot alternative routes between two cities. In each of the two multiple alignments, the two cities, Beijing and New York in this example, are shown as a New pattern in column 0. Each of the remaining columns in each multiple alignment shows an Old pattern, drawn from a pool of such patterns, showing a one-stop flying route between another pair of cities.

\subsection{Theoretical Foundations}\label{dlnn_theoretical_foundations_section}

Several things point to deep problems with NNs---theoretical foundations that are weak or absent. These include: doubts about their biological validity (Section \ref{dlnn_biological_validity_section}), failure to account for learning from a single experience (Section \ref{dlnn_one-trial_learning_section}), concerns about the computational resources required by NNs (Section \ref{dlnn_speed_of_learning_section}), failures in recognition (Section \ref{dlnn_easily_fooled_section}), weaknesses in under-generalisation and over-generalisation (Section \ref{dlnn_under_over_generalisation_section}), problems of transparency in the representation and processing of knowledge (Section \ref{dlnn_transparency_section}), problems with the processing of natural language (Section \ref{dlnn_nl_processing_section}), and problems in modelling reasoning and other `symbolic' kinds of processing (Section \ref{dlnn_symbolic_ai_section}).

That there are deficiencies in theory for NNs is also suggested by their many different versions: `time-delay' neural networks, `gradient-based deep learners with alternating convolutional and down-sampling layers', `weight-sharing feed-forward', `nonlinear auto-regressive with exogenous inputs recurrent', `max-pooling convolutional', `multi-column GPU max-pooling convolutional', `bi-directional long short-term memory recurrent', and more \cite{schmidhuber_2015}. Although these many versions indicate high levels of interest in NNs, they are also an indication that researchers are still not entirely happy with the NN framework.

\section{Unified Theories of Cognition and Related Research}\label{utc_agi_section}

As mentioned in Appendix \ref{occams_razor_appendix}, `unified theories of cognition' and `artificial general intelligence' are honourable exceptions to the rather marked tendency for research in AI and in human perception and cognition to have become fragmented into many subfields, without much communication between them. This section will not attempt to review all the projects in these and related areas. It will say something briefly about two of them: Soar and CogPrime.

The first of these, Soar \cite{laird_2012}, was originally developed as a response to Allen Newell's call for `unified theories of cognition', in his book of the same name \cite{newell_1990}. Up to and including Soar 8 ({\em ca.} 2004), ``...~there has been a single framework for all tasks and subtasks (problem spaces), a single representation of permanent knowledge (productions), a single representation of temporary knowledge (objects with attributes and values), a single mechanism for generating goals (automatic subgoaling), and a single learning mechanism (chunking)'', in accordance with the principle that ``...~the number of distinct architectural mechanisms should be minimized.''\footnote{From the home page of the official Soar website at \href{http://bit.ly/1NXnGP9}{bit.ly/1NXnGP9}, retrieved 2015-09-02.}

But in later versions of Soar, the stress on ``unified'' seems to have diminished: ``We have revisited [the assumption that architectural mechanisms should be minimised] as we attempt to ensure that all available knowledge can be captured at runtime without disrupting task performance. This is leading to multiple learning mechanisms (chunking, reinforcement learning, episodic learning, and semantic learning), and multiple representations of long-term knowledge (productions for procedural knowledge, semantic memory, and episodic memory)'' ({\em ibid.}).

Although Newell uses the expression ``cognitive architecture'' \cite[p.~xii and elsewhere]{newell_1990}, an increased willingness now for Soar to encompass multiple structures and multiple processes appears to be the reason for an increased emphasis on the concept of a cognitive architecture rather than that of a unified theory of cognition.

The variety of structures and mechanisms in the early Soar, and their expansion in the later Soar, suggests that, despite the commendable early quest for a unified model of cognition, the system has been and is a kluge (Appendix \ref{kluge_appendix}). It appears that the SP system scores better in terms of simplicity and power (Appendix \ref{sp_simplicity_power_evaluation_appendix}), doing at least as much but with significantly less structure or mechanism.

Similar things may be said about CogPrime \cite{goertzel_2012}. To be fair, it does not aspire to be a unified theory of cognition or intelligence, since ``We have not discovered any one algorithm or approach capable of yielding the emergence of [intelligence].'' ({\em ibid.}, p.~1). Rather, it is an ``architecture'' for artificial general intelligence containing a variety of structures and mechanisms, with ``{\em cognitive synergy}: the fitting together of different intelligent components into an appropriate cognitive architecture, in such a way that the components richly and dynamically support and assist each other ....'' (p.~2, emphasis in the original), based on the hypothesis that such synergy ``can yield robust intelligence at the human level and ultimately beyond.'' (p.~3).

In effect, the argument is that, although CogPrime is a kluge, ``cognitive synergy'', also described as ``secret sauce'' (p.~2), will somehow overcome that handicap. Although \cite{goertzel_2012} contains outline examples of how cognitive synergy may be applied, it is difficult to see how, via such applications, CogPrime's relatively complex collection of structures and mechanisms might yield ``robust intelligence'', or achieve the overall simplicity that one would wish to see in any good theory.

In \cite{goertzel_2012}, there is a welcome recognition of the importance of Occam's Razor, which is described as ``an overall design principle that underlies nearly every part of the [CogPrime] system.'' (p.~36). But, from an SP perspective, this falls far short of what is needed: it does not recognise the importance of ICMUP (Section \ref{ovv_icmup_section}) or multiple alignment (Section \ref{ovv_multiple_alignment_section}), and it does not recognise how, via those two principles, one may achieve an overall simplification in theory (Section \ref{ovv_simplification_integration_concepts_section}), and an overall simplification in computing systems (Section \ref{ovv_simplification_integration_computers_section}), cutting out much of the complexity that arises when a collection of diverse concepts and mechanisms, from different areas, are bolted together, with or without ``secret sauce''.

It seems that researchers in AI have largely given up on the quest for a truly unified theory of cognition. The nearest thing seems to be a relatively large number of `cognitive architectures' (many of which, up to 2010, are reviewed in \cite{samsonovich_2010}). It appears that the SP system is unique in its attempt to simplify and integrate concepts across a broad canvass. As such, it appears to be unusually strong in combining conceptual simplicity with descriptive or explanatory power (Section \ref{ovv_simplification_integration_concepts_section}).

\section{Universal Search}\label{universal_search_section}

Some ideas, that may be grouped together loosely under the heading `universal search', seem, at first sight, to offer comprehensive solutions to problems in AI and beyond.

Solomonoff \cite{solomonoff_1986} has argued that the great majority of problems in science and mathematics may be seen as either `machine inversion' problems or `time limited optimization' problems, and that both kinds of problem can be solved by inductive inference using the principle of minimum length encoding.

In `Levin search' \cite{levin_1973,levin_1984}, which aims to solve inversion problems, the system is designed to solve any given problem by interleaving all possible programs on a universal Turing machine, sharing computation time equally among them, until one of the executed programs manages to solve the target problem.

Ideas of this kind have been developed by Hutter (eg, \cite{hutter_2002}), Schmidhuber (eg, \cite{schmidhuber_2009a}), and others.\footnote{For a useful overview, see ``Universal search'' by Matteo Gagliolo \cite{gagliolo_2007}.} From the perspective of the SP research programme, the main sources of concern are:

\begin{itemize}

\item The apparent difficulty of translating the abstract concepts of universal search into working models that exhibit aspects of intelligence or are potentially useful.

\item The apparent difficulty of squeezing the subtlety and complexity of human intelligence into the procrustean bed of `well defined problems' \cite{hutter_2002}---something that appears to be a prerequisite for universal search.

\item With problems in AI, it is rarely possible to guarantee solutions that are theoretically ideal---a focus of interest in research on universal search. Normally, via heuristic search, we should aim for solutions that are `reasonably good' and not necessarily perfect (Appendix \ref{heuristic_search_appendix}).

\item In terms of the trade-off between simplicity and power (Appendix \ref{occams_razor_appendix}), it appears that theories in the area of universal search are running the risk of being too simple and over-general, and correspondingly weak in terms of descriptive or explanatory power.

    By contrast, the SP system provides mechanisms for finding good full and partial matches between patterns (Appendix \ref{icmup_appendix}), for building multiple alignments (Appendix \ref{ma_appendix}), and for creating grammars (Appendix \ref{ic_unsupervised_learning_appendix}), and it has plenty to say about a range of observations and concepts in AI and beyond.

\end{itemize}

\section{Bayesian Networks and Some Other Models for AI}\label{models_for_ai_section}

This section considers briefly some other systems or techniques that have been proposed as models for AI, or aspects of AI: Bayesian networks, support vector machines, hidden Markov models, Kalman filters, self-organising maps, Petri nets, cellular automata, pushdown automata, dimensionality reduction, and finite-state automata. It is likely that there are other systems to which the remarks in this section apply.

The suggestion here, which is admittedly a rather sweeping generalisation, is that, while these models are admirably simple, and while they may have useful applications, they lack the descriptive and explanatory range of the SP system. In terms of the quest for a favourable combination of simplicity and power (Appendix \ref{occams_razor_appendix} and Section \ref{ovv_simplification_integration_concepts_section}), their descriptive and explanatory range is too limited.

With appropriate data, the SP system provides an alternative to Bayesian networks in modelling such phenomena as `explaining away' (\cite[Section 10.2]{sp_extended_overview}, \cite[Section 7.8]{wolff_2006}). In addition to its relative strength in terms of simplicity and power, the SP system, compared with a Bayesian network, has three apparent advantages:

\begin{itemize}

\item {\em Simplicity in the representation of statistical knowledge}. Each node in a Bayesian network contains a table of conditional probabilities for possible combinations of inputs, and these tables can be quite large. By contrast, the SP framework only requires a single measure of frequency for each pattern. Absolute or conditional probabilities can be derived from that frequency measure, as required.

\item {\em Objectivity}. Arguably, the SP system has an advantage compared with Bayesian networks because it derives `objective' probabilities in a straightforward way from frequencies of occurrence, and it eliminates the subjectivity in Bayes' theorem in its concept of probability as a degree of belief.

\item {\em Creation of ontologies}. Bayes' theorem assumes that the categories that are to be related to each other via conditional probabilities are already established. By contrast, the SP system provides an account of how a knowledge of categories and entities may be developed via unsupervised learning (Appendix \ref{ic_unsupervised_learning_appendix}).

\end{itemize}

\section{IBM's Watson}\label{dlnn_watson_section}

As is now well known, a team of researchers at IBM developed a computing system, called Watson, that, in 2011, beat the best human players at the TV quiz game {\em Jeopardy!}\footnote{See, for example, ``Watson (computer)'', {\em Wikipedia}, retrieved 2015-08-12, \href{http://bit.ly/1DwVKiC}{bit.ly/1DwVKiC}.}

Of course winning at {\em Jeopardy!} is a major achievement, with potential benefits in terms of ideas and, perhaps, applications. But there are concerns about aspects of Watson and its development, described in the following subsections.

\subsection{Significance of Watson for AI}\label{watson_significance_for_ai_section}

 Doubt has been expressed about the significance of Watson for AI:

\begin{quote}

    ``...~systems that seem to have mastered complex language tasks, such as IBM's {\em Jeopardy!} winner Watson, do it by being super-specialized to a particular format. `It's cute as a demonstration, but not work that would really translate to any other situation,' [says Yann LeCun].'' \cite[p.~73]{simonite_2015}.

\end{quote}

\subsection{Ignoring Research on Human Perception and Cognition}\label{ignoring_cognition_research_section}

There is concern about how, in the development of Watson, research on human perception and cognition was ignored. Dave Ferrucci, leader of the team that developed Watson, has been quoted as saying:

    \begin{quote}

        ``Did we sit down when we built Watson and try to model human cognition? Absolutely not. We just tried to create a machine that could win at {\em Jeopardy!}''\footnote{``The man who would teach machines to think'', {\em The Atlantic}, November 2013, \href{http://theatln.tc/1fi9AFv}{theatln.tc/1fi9AFv}.}

    \end{quote}

An implication of this remark is that, by ignoring the large body of research into human cognition and the many insights that have been gained, the developers of Watson have probably reduced the chances of Watson providing a meaningful route to `cognitive computing' as described in \cite{kelly_hamm_2013}.

\subsection{Watson as a Kluge}\label{watson_as_kluge_section}

The original Watson was created by combining natural language understanding with statistical analysis of very large amounts of text. IBM has now added capabilities in translation, speech-to-text, and text-to-speech, and they plan to add capabilities in deep learning,\footnote{``IBM pushes deep learning with a Watson upgrade'', {\em MIT Technology Review}, 2015-07-09, \href{http://bit.ly/1Nq0bMg}{bit.ly/1Nq0bMg}.} with large numbers of medical images as data for learning.\footnote{``Why IBM just bought billions of medical images for Watson to look at'', {\em MIT Technology Review}, 2015-08-11, \href{http://bit.ly/1P6lcvQ}{bit.ly/1P6lcvQ}.}

Further, Watson was created as a ``holistic combination of many diverse algorithmic techniques'' \cite[p.~2]{ferrucci_2012} or ``hundreds of different cooperating algorithms'' \cite[pp.~2--3]{ferrucci_2012} which are, as Ferrucci remarks, reminiscent of Minsky's {\em Society of Mind} \cite{minsky_1986}.

In short, Watson was originally developed, and is continuing to be developed, as a kluge: ``a clumsy or inelegant---yet surprisingly effective---solution to a problem.'' \cite[p.~2]{marcus_2008}.

Does this matter? There are several possible answers to this question. On the plus side (the kluge is a good thing):

\begin{itemize}

\item It is probably possible to create a system that does useful things and earns money.

\item Combining technologies may help to overcome weaknesses in individual technologies and it may help to overcome fragmentation in AI (``If deep learning can be combined with other AI techniques effectively, that could produce more rounded, useful systems.''\footnote{``IBM pushes deep learning with a Watson upgrade'', {\em MIT Technology Review}, 2015-07-09, \href{http://bit.ly/1Nq0bMg}{bit.ly/1Nq0bMg}.}).

\item Since the human mind is a kluge \cite{marcus_2008}, it should not matter if AI systems are the same (but see Appendix \ref{kluge_appendix}).

\end{itemize}

\noindent But on the minus side:

\begin{itemize}

\item Creating a kluge may yield short-term gains but is unlikely to be satisfactory on longer timescales \cite[Sections 2, 6, and 7]{sp_benefits_apps}.

\item Creating a kluge may be a distraction from the long-term goal of developing `cognitive computing':

    \begin{quote}

        ``The creation of this new era of [cognitive] computing is a monumental endeavor ...~no company can take on this challenge alone. So we look to our clients, university researchers, government policy makers, industry partners, and entrepreneurs---indeed the entire tech industry---to take this journey with us.'' \cite[Preface]{kelly_hamm_2013}.

    \end{quote}

\end{itemize}

On balance, a two-pronged strategy is probably best: take advantage of short-term gains that may accrue from developing applications such as Watson as kluges and, at the same time, develop cleaner and more elegant solutions to problems in AI, drawing on insights gained from kluges (Appendix \ref{long_short_perspectives_appendix}). The SP system is a good candidate for inclusion in the second strand of research.

\subsection{Watson and Energy Consumption}\label{watson_energy_consumption_section}

Apart from the concerns outlined above, a major objection is that Watson does nothing to solve the problem of energy consumption in the processing of big data, outlined in Section \ref{computational_energy_efficiency_of_computers_section}. Indeed, the projected addition of deep learning to Watson, mentioned in Section \ref{watson_as_kluge_section}, is likely to make things worse owing to inefficiencies with deep learning and artificial neural networks, outlined in Section \ref{dlnn_speed_of_learning_section}.

\subsection{A Possible Contribution From the SP System}

In keeping with the two-pronged strategy mention in Section \ref{watson_as_kluge_section}, there is probably a case for recasting Watson in the SP framework. There is potential to create a system that combines the strengths of Watson with the strengths of the SP system: in aspects of AI, in its origins in research on human perception and cognition, in its potential to simplify and integrate diverse kinds of knowledge and diverse kinds of processing, and in its potential for very substantial cuts in energy consumption.

\section{Big Data and Autonomous Robots}\label{big_data_autonomous_robots_section}

Potential benefits and applications for the SP theory are summarised in Appendix \ref{benefits_applications_appendix}. This section gives a bit more detail about two areas of potential application: big data and autonomous robots.

\subsection{Big Data}\label{big_data_section}

The paper ``Big data and the SP theory of intelligence'' \cite{sp_big_data} describes how the SP theory may help to solve nine problems with big data:

\begin{itemize}

\item {\em Helping to overcome the problem of variety in big data}. The SP system may serve as a universal framework for the representation and processing of knowledge (UFK), helping to tame the great variety of formalisms and formats for data, each with its own mode of processing (Section \ref{representation_processing_knowledge_section}).

\item {\em Learning and discovery}. In accordance with the DONSVIC principle (Section \ref{ovv_donsvic_section}, \cite[Section 5.2]{sp_extended_overview}), the system has strengths in the unsupervised learning or discovery of `natural' structures in data, with potential for further development.

\item {\em Interpretation of data}. The SP system has strengths in areas such as pattern recognition, information retrieval, parsing and production of natural language, translation from one representation to another, several kinds of reasoning, planning and problem solving.

\item {\em Velocity: analysis of streaming data}. The SP system lends itself to an incremental style, assimilating information as it is received, much as people do.

\item {\em Volume: making big data smaller}. Reducing the size of big data via lossless compression can yield direct benefits in the storage, management, and transmission of data, and indirect benefits in several of the other areas outlined in this subsection and discussed in \cite{sp_big_data}.

\item {\em Supercharging the transmission of data}. In addition to economies in the transmission of data via simple reductions in volume, there is potential for additional and very substantial economies in the transmission of data by judicious separation of `encoding' and `grammar'.

\item {\em Computational and energy efficiency}. There is potential for large gains in the computational efficiency of computers, with corresponding savings in the use of energy in computing, and for reductions in the size and weight of computers (Section \ref{computational_energy_efficiency_of_computers_section}).

\item {\em Veracity: managing errors and uncertainties in data}. The SP system can identify possible errors or uncertainties in data, suggest possible corrections or interpolations, and calculate associated probabilities.

\item {\em Visualisation}. Knowledge structures created by the system, and inferential processes in the system, are all transparent and open to inspection. They lend themselves to display with static and moving images.

\end{itemize}

Considering these proposed solutions collectively, and in several cases individually, it appears that there are no alternatives that can rival the potential of what is described in \cite{sp_big_data}.

\subsection{Autonomous Robots}\label{autonomous_robots_section}

The paper ``Autonomous robots and the SP theory of intelligence'' \cite{sp_autonomous_robots} describes how the SP theory may help in the design of the information-processing `brains' of autonomous robots:

\begin{itemize}

\item {\em Computational and energy efficiency}. This is a revised version of the discussion in \cite[Section IX]{sp_big_data}.

\item {\em Towards human-like versatility in intelligence}. The strengths of the SP system in diverse areas, summarised in Appendix \ref{empirical_conceptual_support_appendix}, can help in the development of human-like versatility in autonomous robots.

\item {\em Towards human-like adaptability in intelligence}. It appears that unsupervised learning in the SP framework has potential as a key to human-like adaptability in intelligence, both directly and as a basis for other kinds of learning.

\end{itemize}

This approach to the development of intelligence in autonomous robots is quite different from others, and arguably more promising.

\section{Distinctive Features and Advantages of the SP System in Comparison With Some Symbolic Alternatives}\label{other_alternatives_section}

This section summarises distinctive features and advantages of the SP system in symbolic kinds of computing. {\em A key advantage of the SP system in these areas (and others discussed elsewhere in this paper) is that it provides for the seamless integration of each function with other aspects of intelligence} (Section \ref{ovv_simplification_integration_computers_section}). This applies in each of the subsections below but will not be constantly repeated.

\subsection{Pattern Recognition and Vision}\label{pattern_recognition_vision_section}

The main strengths and potential of the SP system in pattern recognition are described in \cite[Chapter 6]{wolff_2006} and \cite[Section 9]{sp_extended_overview}. In brief, these are: recognition at multiple levels of abstraction (Figure \ref{class-inclusion_figure}) and via hierarchies of parts and sub-parts (Figure \ref{part_whole_figure}); recognition of `family resemblance' or polythetic categories; recognition that is robust in the face of errors of omission, commission or substitution (Section \ref{pattern_recognition_vision_section}); for each act of recognition, calculation of associated probabilities; modelling the way in which context may influence recognition.

In addition to the strengths just mentioned, there is, in computer vision \cite{sp_vision}, potential in the SP system in: scene analysis via the parsing of an image; the learning of visual entities and classes of entity; the creation of 3D models of objects and of their surroundings; explaining how we can see things that are not objectively present (Section \ref{amodal_perception_completion_section}); providing insights into the phenomena of size constancy, lightness constancy, and colour constancy; and finding alternative interpretations of images and scenes.

The two subsections that follow amplify some of the points above.

\subsubsection{Recognition in the Face of Errors of Omission, Commission, and Substitution}\label{recognition_with_errors_section}

\begin{figure}[!htbp]
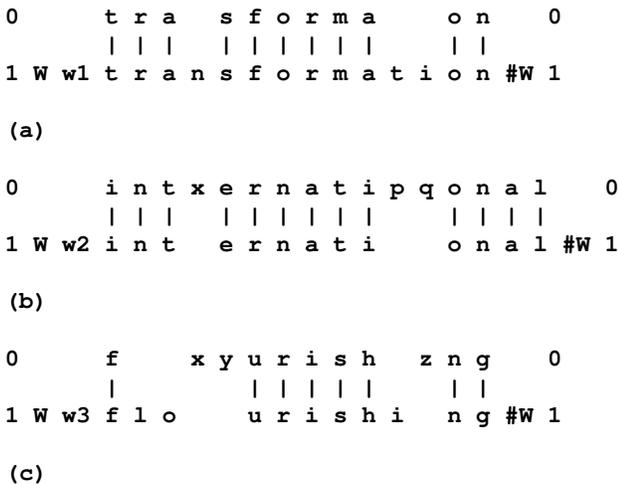

\fontsize{09.00pt}{10.80pt}
\centering
{\bf
\begin{BVerbatim}
0      t r a   s f o r m a     o n    0
       | | |   | | | | | |     | |
1 W w1 t r a n s f o r m a t i o n #W 1

(a)

0      i n t x e r n a t i p q o n a l    0
       | | |   | | | | | |     | | | |
1 W w2 i n t   e r n a t i     o n a l #W 1

(b)

0      f     x y u r i s h   z n g    0
       |         | | | | |     | |
1 W w3 f l o     u r i s h i   n g #W 1

(c)
\end{BVerbatim}
}
\caption{Three multiple alignment illustrating recognition with errors of omission, commission, and substitution, as described in the text.}% Reproduced with permission from \cite[Figure 6]{sp_vision}.
\label{errors_figure}
\end{figure}

Figure \ref{errors_figure} shows how, via the creation of multiple alignments, the SP system may recognise words despite errors of omission ((a) in the figure), commission (b), and substitution (c).\footnote{By contrast with most multiple alignments shown in this paper, the ones in this figure and also in Figures \ref{occlusion_figure} and \ref{learning_alignment_figure} have been rotated by $90\degree$, replacing columns with rows. The choice between these two representations, which are equivalent, depends on what fits best on the page.}

These examples are very simple but this capability of the SP system applies in much more complex examples of pattern recognition and also in other areas such as the processing of natural language (Section \ref{nlp_section}).

\subsubsection{Amodal Perception and Completion}\label{amodal_perception_completion_section}

\begin{quote}

``A remarkable property of human perception is the ease with which our visual system interpolates information not directly visible in an image. A particularly prominent example of this ...~is {\em amodal perception}: the phenomenon of perceiving the whole of a physical structure when only a portion of it is visible'' (\cite[pp.~1--2]{zhu_etal_2015}, emphasis in the original).

\end{quote}

A familiar example of amodal perception is how we can recognise something like a vase with flowers even when it is partly hidden or `occluded' by something like a mug. Here, we have no difficulty with `amodal completion': imagining the parts of the vase and flowers that we would see if the mug was removed.

\begin{figure}[!htbp]
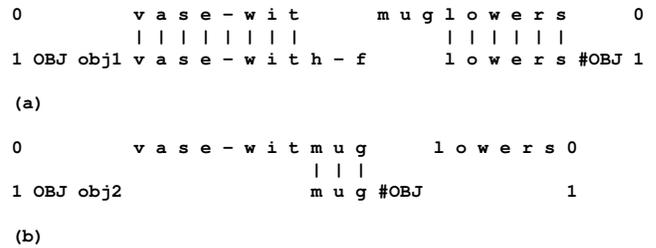

\fontsize{07.00pt}{08.40pt}
\centering
{\bf
\begin{BVerbatim}
0          v a s e - w i t       m u g l o w e r s      0
           | | | | | | | |             | | | | | |
1 OBJ obj1 v a s e - w i t h - f       l o w e r s #OBJ 1

(a)

0          v a s e - w i t m u g      l o w e r s 0
                           | | |
1 OBJ obj2                 m u g #OBJ             1

(b)
\end{BVerbatim}
}
\caption{Two multiple alignments modelling perception with occlusion, as described in the text.}
\label{occlusion_figure}
\end{figure}

\sloppy Figure \ref{occlusion_figure} shows how the SP computer model may solve a one-dimensional analogue of this perceptual problem. Row 0 in each of the two multiple alignments contains the pattern `\texttt{v a s e - w i t m u g l o w e r s}', which is an analogue of a vase with flowers which is partly obscured by a mug. The first multiple alignment shows how the vase with flowers---represented by the pattern `\texttt{OBJ obj1 v a s e - w i t h - f l o w e r s \#OBJ}' in row 1---may be recognised despite its partial occlusion by `\texttt{m u g}', and how the missing characters (`\texttt{h - f}') may be interpolated. The second multiple alignment shows how the mug (represented by the pattern `\texttt{OBJ obj2 m u g \#OBJ}' in row 1) may be recognised as a separate entity.

This example does not model the way we perceive the mug to lie in front of the vase with flowers. But, as described in \cite[Sections 6.1 and 6.2]{sp_vision}, the SP system has potential for the building of 3D models. If that potential can be realised, the capability may potentially be married with amodal perception of one object that is partly obscured by another.

\begin{figure}[!htbp]
\centering
\includegraphics[width=0.3\textwidth]{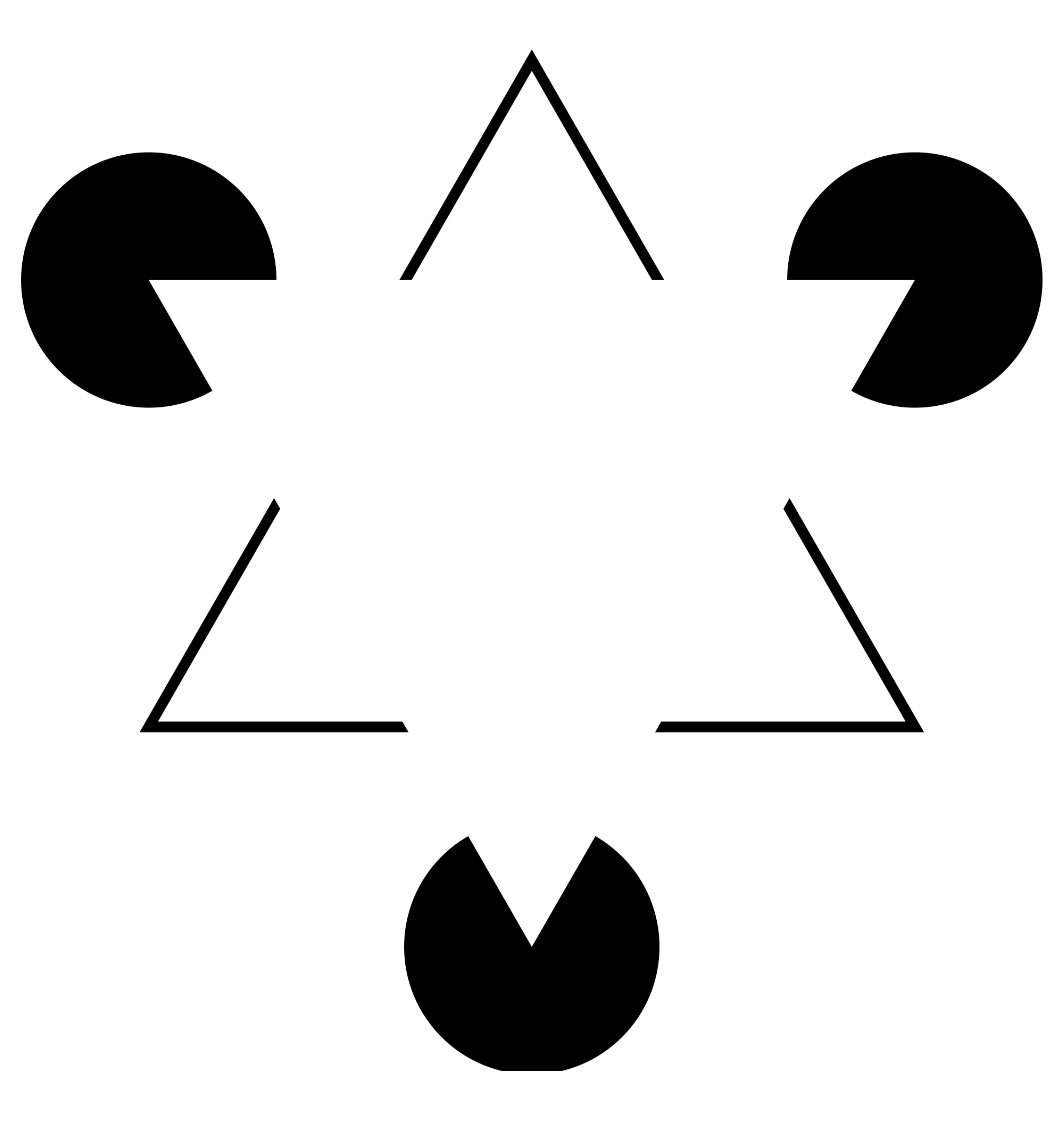}
\caption{Kanizsa's triangle.}
\label{kanizsas_triangle_figure}
\end{figure}

A popular example of amodal completion is Kanizsa's triangle, shown in Figure \ref{kanizsas_triangle_figure}. Here, we `see' a white equilateral triangle in the middle of the figure with its apex at the bottom, although it is marked only with a minimum of features: three corners, each in a black disk, and breaks in the sides of another equilateral triangle with its apex at the top.

\begin{figure*}[!htbp]
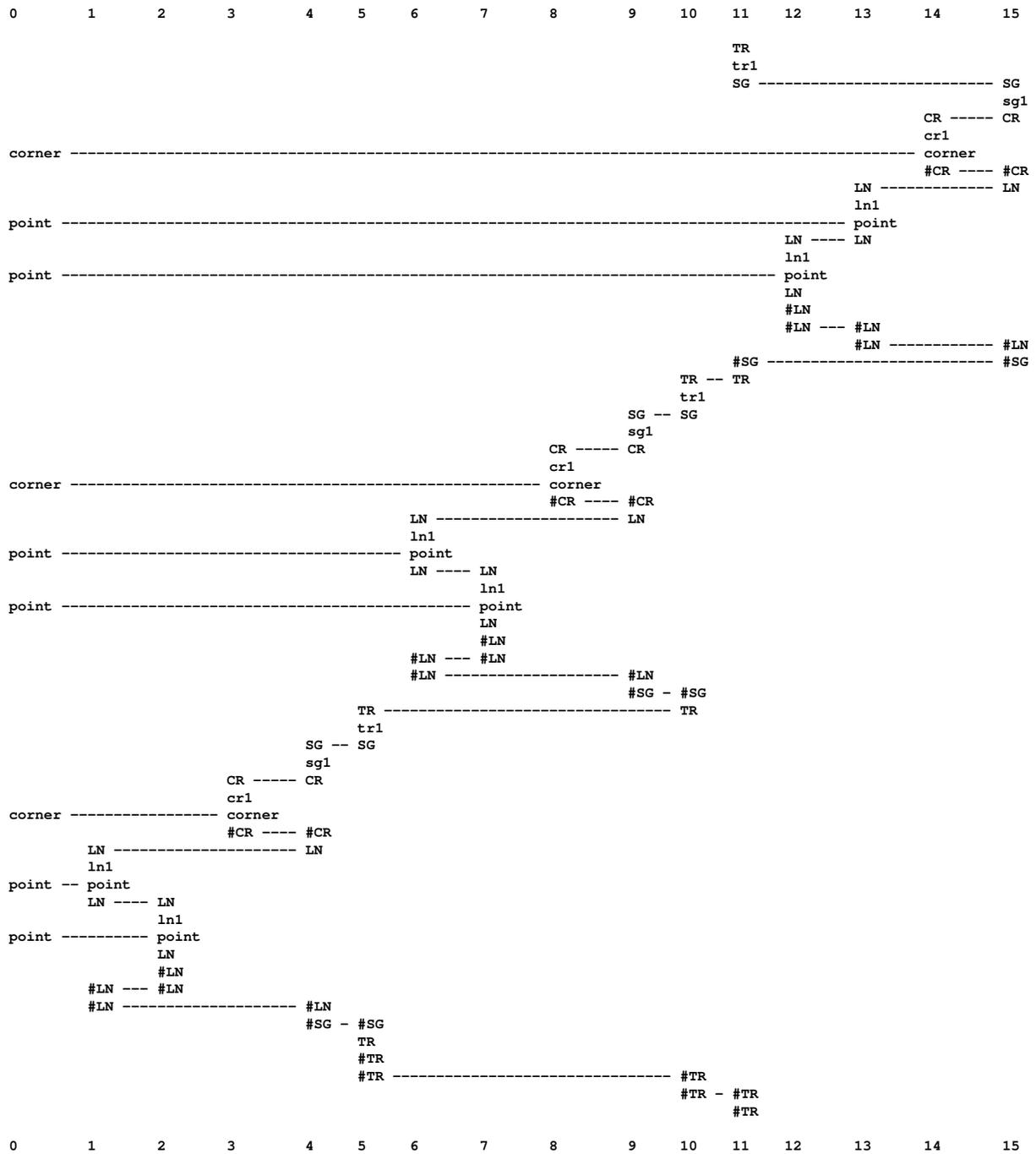

\fontsize{06.50pt}{07.80pt}
\centering
{\bf
\begin{BVerbatim}
0        1       2       3        4     5     6       7       8        9     10    11    12      13      14       15

                                                                                   TR
                                                                                   tr1
                                                                                   SG --------------------------- SG
                                                                                                                  sg1
                                                                                                         CR ----- CR
                                                                                                         cr1
corner ------------------------------------------------------------------------------------------------- corner
                                                                                                         #CR ---- #CR
                                                                                                 LN ------------- LN
                                                                                                 ln1
point ------------------------------------------------------------------------------------------ point
                                                                                         LN ---- LN
                                                                                         ln1
point ---------------------------------------------------------------------------------- point
                                                                                         LN
                                                                                         #LN
                                                                                         #LN --- #LN
                                                                                                 #LN ------------ #LN
                                                                                   #SG -------------------------- #SG
                                                                             TR -- TR
                                                                             tr1
                                                                       SG -- SG
                                                                       sg1
                                                              CR ----- CR
                                                              cr1
corner ------------------------------------------------------ corner
                                                              #CR ---- #CR
                                              LN --------------------- LN
                                              ln1
point --------------------------------------- point
                                              LN ---- LN
                                                      ln1
point ----------------------------------------------- point
                                                      LN
                                                      #LN
                                              #LN --- #LN
                                              #LN -------------------- #LN
                                                                       #SG - #SG
                                        TR --------------------------------- TR
                                        tr1
                                  SG -- SG
                                  sg1
                         CR ----- CR
                         cr1
corner ----------------- corner
                         #CR ---- #CR
         LN --------------------- LN
         ln1
point -- point
         LN ---- LN
                 ln1
point ---------- point
                 LN
                 #LN
         #LN --- #LN
         #LN -------------------- #LN
                                  #SG - #SG
                                        TR
                                        #TR
                                        #TR -------------------------------- #TR
                                                                             #TR - #TR
                                                                                   #TR

0        1       2       3        4     5     6       7       8        9     10    11    12      13      14       15
\end{BVerbatim}
}
\caption{A multiple alignment modelling a one-dimensional analogue of how Kanizsa's triangle may be perceived, as described in the text.}
\label{kt_multiple_alignment_figure}
\end{figure*}

Figure \ref{kt_multiple_alignment_figure} shows how the SP computer model may recognise a one-dimensional analogue of such a figure via the building of a multiple alignment, using the features that are available. Column 0 shows a New pattern representing the minimal features: each `\texttt{corner}' in the multiple multiple alignment represents a corner in a black disk in Figure \ref{kanizsas_triangle_figure}, and each `\texttt{point}' in the multiple alignment is a point on a side of the target triangle where there is break in the side of the other triangle.

In this example, recognition is achieved via the parsing of the New pattern in terms of Old patterns representing elements of the triangle: the pattern `\texttt{TR}' in column 11 represents the whole triangle, `\texttt{SG}' in columns 4, 9, and 15 represents a segment of the triangle comprising a corner (`\texttt{CR}') and a line (`\texttt{LN}') which is itself a sequence of points.

In this example, there is recursion (like that shown in Figure \ref{recursion_figure}) at two levels: the triangle is a recursive grouping of segments (so that, strictly speaking, it represents a polygon, not a triangle), and the line is built from a recursive sequence of points.

The example illustrates amodal completion because a mere two points on each side of the Kanizsa triangle are cues for what is potentially an infinite sequence of points between one corner and another.

\subsection{The Processing of Natural Language}\label{nlp_section}

With the processing of natural language, the main strengths and potential of the SP system are described in \cite[Chapter 5]{wolff_2006} and \cite[Section 8]{sp_extended_overview}. In brief: it can model the parsing of natural language directly and transparently, as illustrated in Figure \ref{parsing_figure}; it provides a simple, direct means of encoding discontinuous dependencies in syntax; in parsing, it can accommodate syntactic ambiguities, and the resolution of ambiguities via the provision of appropriate context; parsing may be achieved via recursive structures in syntax (Figure \ref{recursion_figure}); parsing is robust against errors of omission, commission, or substitution (much as in pattern recognition---Section \ref{pattern_recognition_vision_section}); and one mechanism may achieve both the parsing and production of natural language.

A key advantage of the SP system compared with most other systems for the processing of natural language is that, because all kinds of knowledge---both syntactic and semantic---are represented in one simple format (SP patterns) and because all kinds of knowledge are processed in the same way (via the formation and processing of multiple alignments), the system has clear potential to facilitate the seamless integration of syntax and semantics. In a similar vein, the processing of natural language may be integrated seamlessly with other aspects of intelligence.

\subsection{Unsupervised Learning of Natural Language}\label{nl_learning_section}

Although grammatical inference has been the subject of research for many years, automatic learning of the syntax of a natural language remains a major challenge. Even more difficult is unsupervised learning of the rich structure of non-linguistic knowledge which provides the `meanings' for language, and unsupervised learning of the kinds of syntactic-semantic structures that are needed for such things as interpreting the meaning of natural language, the production of speech or writing from meanings, and, when it is done at a high standard, translation from one language to another (Section \ref{dlnn_nl_translation_section}).

As mentioned in Appendix \ref{outline_of_sp_theory_appendix}, the SP programme of research grew out of earlier research developing computer models of language learning, but it has required a radical reorganisation of earlier models to meet the new goals of the programme. Now, the SP computer model demonstrates unsupervised learning of plausible generative grammars for the syntax of English-like artificial languages, including the learning of segmental structures, classes of structure, and abstract patterns \cite[Chapter 9]{wolff_2006}, in accordance with the DONSVIC principle (Section \ref{ovv_donsvic_section}).

A key insight from this research, discussed briefly in Section \ref{dlnn_generalisation_compression_section}, is that information compression appears to solve the problems of both over- and under-generalisation in the learning of language (\cite[Section 9.5.3]{wolff_2006}, \cite[Section 5.3]{sp_extended_overview}). This appears to be a significant advantage compared with, for example, deep learning with neural networks (Section \ref{dlnn_generalisation_compression_section}).

It appears that, with some further work, the potential is considerable: in the learning of syntactic structures, in the learning of semantic structures, and in the learning of syntactic-semantic structures. The potential of the SP system in these areas appears to be a major advantage of the system compared with alternatives.

\subsection{Exact and Inexact Forms of Reasoning}\label{exact_inexact_reasoning_section}

From the General Problem Solver \cite{newell_simon_1972}, through Prolog \cite{clocksin_mellish_2003}, to such systems as Description Logics \cite{horrocks_2008}, concepts derived from classical logic have been prominent in AI and related fields such as the semantic web.

Although the all-or-nothing certainties of classical logic can be useful, it has been recognised for some time that much of human thinking and reasoning revolves around judgements that may have varying levels of certainty. This has led to several proposals for systems that, in one way or another, combine exact or `logical' reasoning with inexact, `fuzzy' or `probabilistic' kinds of reasoning (eg, \cite{zadeh_1965,klinov_parsia_2013}). Some recent developments are described in \cite{russell_2015}.

As described in \cite[Chapter 7]{wolff_2006}, a distinctive feature and apparent advantage of the SP system is that it is fundamentally probabilistic (Appendix \ref{ic_prediction_probabilities_appendix}) but levels of confidence may be increased by gathering more evidence, or by concentrating on probabilities that are close to 0 or 1, or both those things. And the SP system can model several of the concepts that are familiar in logic and mathematics (such as {\em variable}, {\em value}, and {\em type}), as described in \cite[Chapter 10]{wolff_2006} and \cite[Section 6.6]{sp_benefits_apps}. There is clear potential for the system to model both probabilistic and exact forms of reasoning and to switch between them according to need.

The SP system also demonstrates several more specific kinds of reasoning within one unified framework (\cite[Section 6.4, Chapters 7 and 10]{wolff_2006}, \cite[Section 10]{sp_extended_overview}): one-step `deductive' reasoning; abductive reasoning; reasoning with probabilistic decision networks and decision trees; reasoning with `rules'; nonmonotonic reasoning and reasoning with default values; reasoning in Bayesian networks, including `explaining away' (Section \ref{models_for_ai_section}); causal diagnosis; reasoning which is not supported by evidence; and inheritance of attributes in an object-oriented class hierarchy or heterarchy. There is also potential for spatial reasoning \cite[Section IV-F.1]{sp_autonomous_robots} and what-if reasoning \cite[Section IV-F.2]{sp_autonomous_robots}.

Because these several kinds of reasoning all flow from one unified framework, they may be used together in any combination according to need. In the same vein, these several forms of reasoning may integrate seamlessly with other aspects of intelligence: pattern recognition, natural language processing, unsupervised learning, and so on. This kind of flexibility is a distinctive feature and apparent advantage of the SP system compared with alternatives (Section \ref{ovv_simplification_integration_computers_section}).

\subsection{Representation and Processing of Diverse Forms of Knowledge}\label{representation_processing_knowledge_section}

A problem with AI and other areas of computing as they have developed to date is that knowledge may be represented with a large number of different formalisms, and often, for each one, there is a large number of different formats. This complexity is compounded by the fact that, normally, each formalism and each format has its own mode of processing. Until recently, this complexity has been easy to ignore. But with the advent of big data, it has become a major problem, a problem that the SP system may help to solve (Section \ref{big_data_section}).

As described in outline in \cite[Section III-B]{sp_big_data}, and in more detail in \cite{wolff_2006,sp_extended_overview}, the SP system promotes the seamless integration of a wide a variety of kinds of knowledge, with seamless integration of their processing. These include: the syntax of natural language; class hierarchies, part-whole hierarchies, and their integration; trees and networks, including Bayesian networks; entity-relationship structures; relational knowledge (tuples); if-then rules, associations, and other knowledge in support of reasoning; patterns and images; structures in three dimensions; and sequential and parallel procedures.

Since information compression via ICMUP and multiple alignment is at the heart of how knowledge is represented in the SP system, since information compression can in principle be an efficient means of representing any kind of knowledge, and since the multiple alignment framework appears to be a very general means of compressing information, there is reason to believe that {\em any} kind of knowledge---perhaps including the kind of `commonsense' knowledge that is described and discussed in \cite{davis_marcus_2015}---may be represented effectively in the SP system. Similar things may be said, {\em mutatis mutandis}, about reasoning and other kinds of processing of knowledge in the SP system.

\subsection{Software Engineering}\label{software_engineering_section}

Although machine learning is beginning to make an impact, controlling what computers do is still done largely in the traditional manner, via the running of programs and their creation by people. Although machine learning may eventually become dominant, it is likely that, with all kinds of computing system, there will be a continuing need for the foreseeable future, for each system to be controllable by human-created software.

At first sight, the SP system fails this test. SP patterns don't look like a conventional program and the building of multiple alignments does not look much like the running of a conventional program. But, as noted in Section \ref{exact_inexact_reasoning_section}, the multiple alignment framework can model such concepts as {\em variable}, {\em value}, and {\em type}. More generally, the SP system can model the kinds of concepts used in software engineering including {\em procedure}, {\em function} with {\em parameters}, conditional statements, repetition of procedures, the integration of programs and data, and elements of object-oriented design, including {\em class hierarchies} and {\em inheritance} \cite[Section 6.6]{sp_benefits_apps}. The SP system also has potential for the processing of parallel streams of information \cite[Sections V-G, V-H, and V-I, and Appendix C]{sp_autonomous_robots}.

Because of these features of the SP system, and because of its strengths in unsupervised learning (\cite[Chapter 9]{wolff_2006}, \cite[Section 5]{sp_extended_overview}), it has potential for `automatic programming'---the integration of programming with unsupervised learning---as outlined in \cite[Section 6.6.4]{sp_benefits_apps}. This is a distinctive feature of the system and an apparent advantage compared with systems that are specialised only for learning.

\section{Conclusion}

Preceding sections of this paper have aimed to highlight distinctive features of the {\em SP theory of intelligence} and its apparent advantages compared with some AI-related alternatives.

Section \ref{overview_section} summarises distinctive features and strengths of the SP system:

\begin{itemize}

\item Simplification and integration of observations and concepts;

\item Simplification and integration of structures and processes in computing systems;

\item The SP theory is itself a theory of computing;

\item The theory provides the basis for new architectures for computers;

\item Information compression via the matching and unification of patterns is central in the theory;

\item More specifically, all processing is done via the building of {\em multiple alignments}, a concept borrowed and adapted from bioinformatics;

\item Transparency in the representation and processing of knowledge;

\item The unsupervised learning of `natural' structures via information compression (DONSVIC);

\item Interpretation of aspects of mathematics in terms of the SP theory;

\item Interpretation of phenomena in human perception and cognition;

\item Realisation of abstract concepts in terms of neurons and their inter-connections ({\em SP-neural}).

\end{itemize}

In several sections, distinctive features and advantages of the SP system have been highlighted in comparison with AI-related alternatives:

\begin{itemize}

\item The concept of minimum length encoding and related concepts;

\item How computational and energy efficiency in computing may be achieved.

\item Deep learning in neural networks;

\item Unified theories of cognition and related research.

\item Concepts of universal search;

\item Bayesian networks and some other models for AI;

\item IBM's Watson;

\item Solving problems associated with big data;

\item Solving problems in the development of intelligence in autonomous robots.

\item The processing of natural language;

\item Unsupervised learning of natural language;

\item Exact and inexact forms of reasoning;

\item Representation and processing of diverse forms of knowledge;

\item Software engineering.

\end{itemize}

The main conclusion of the paper is that, while some alternatives to the SP system may deliver applications fairly quickly, a strength of the SP system is that it can provide a firm foundation for the long-term development of AI, with many potential benefits and application, and, at the same time, it may deliver useful results on relatively short timescales.

It is envisaged that a high-parallel, open-source version of the SP machine will be created, hosted on an existing high-performance computer, and derived from the existing SP computer model (Appendix \ref{future_developments_appendix}). This would be a means for researchers everywhere to explore what can be done with the system, and to create new versions of it.

\section*{Appendices}

% \appendix

\section{Outline of the SP Theory of Intelligence and SP Machine}\label{outline_of_sp_theory_appendix}

As noted in the Introduction, the {\em SP theory of intelligence} is designed to simplify and integrate observations and concepts across artificial intelligence, mainstream computing, mathematics, and human perception and cognition, with information compression via {\em multiple alignment} as a unifying theme. As outlined in Appendix \ref{sp_computer_model_and_machine_appendix}, the SP theory is realised in the SP computer model which may be regarded as a version of the {\em SP machine}.

The SP theory originates in part from an earlier programme of research on grammatical inference and the unsupervised learning of natural language, with information compression at centre stage \cite{wolff_1988}. However, meeting the goals of the SP research programme has meant a radical reorganisation of the system, with the development of a concept of {\em multiple alignment} (Appendix \ref{ma_appendix}) as a framework for the simplification and integration of diverse structures and functions \cite[Section V-A.4]{sp_autonomous_robots}.

This Appendix is intended to provide readers with sufficient information about the SP system to make the rest of the paper intelligible. The outline starts with a short informal account of the SP system, followed by subsections that describe the main elements of the system in a little more detail.

\subsection{An Informal Account of How the SP System Works}\label{informal_account_of_sp_system_appendix}

\begin{figure}[!htbp]
\centering
\includegraphics[width=0.4\textwidth]{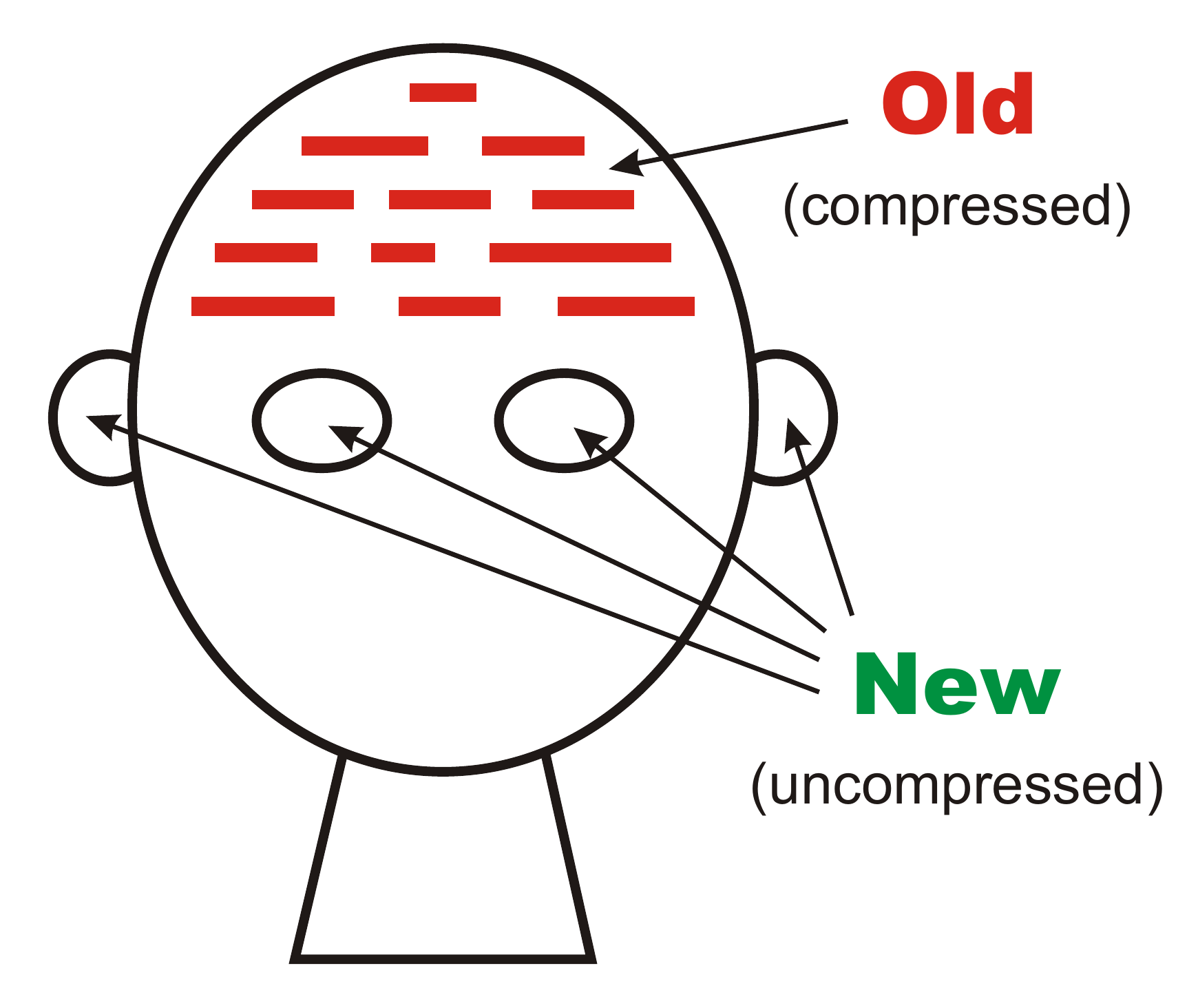}
\caption{Schematic representation of the SP system from an `input' perspective.}
\label{sp_input_perspective_figure}
\end{figure}

The SP theory is conceived as an abstract brain-like system that, in an `input' perspective, may receive {\em New} `patterns' via its senses, and compress some or all of it to create {\em Old} `patterns', as illustrated schematically in Figure \ref{sp_input_perspective_figure}.

In the early stages, when there is little or no Old information in store, the system simply stores New patterns directly as Old patterns, except that `identification' (`ID') symbols are added to each pattern for use later.

After a while, New patterns are received that are fully or partially the same as Old patterns. Then, the system builds multiple alignments as outlined in Appendix \ref{ma_appendix}. From the multiple alignments:

\begin{itemize}

  \item When a New pattern is exactly the same as a stored Old pattern, the system records the occurrence of that pattern in terms of its ID-symbols, not the pattern itself. Since the ID-symbols are normally relatively short compared with their associated pattern, the effect is to store the New pattern in a compressed form.

  \item When a New pattern is received that is partially the same as one or more Old patterns, the system constructs patterns from the parts of the New and Old patterns that match each other and from the parts that do not match each other. Each newly-created pattern is assigned its own ID-symbols and is stored as an Old pattern. The system also constructs `abstract' patterns that, using relevant ID-symbols, record the sequential relationship of the newly-constructed patterns.\footnote{It is envisaged that similar principles will apply when the SP computer model has been developed for the representation and processing of 2D patterns.}

  \item Similar principles apply when the system builds multiple alignments containing previously-constructed abstract patterns.

\end{itemize}

Many of the Old patterns that are created in this way may be seen as `good', but, normally, the system also creates many that people are likely to regard as `bad'. To get rid of the bad patterns, SP system applies a process for evaluating sets of Old patterns (called {\em grammars}) in terms of their ability to compress incoming New information. The one or two grammars that yield high levels of compression are retained by the system while residual patterns---those not found in the successful grammars---are discarded.

It is interesting to see that, very often, patterns that are `good' in terms of information compression are also `good' in terms of human judgements, and {\em vice versa}. This correspondence between what people regard as `natural' and what yields high levels of compression is termed the `DONSVIC' principle (Section \ref{ovv_donsvic_section}).

The building of multiple alignments provides the means by which New information may be encoded economically in terms of Old information. It also provides much of the versatility of the system in such functions as unsupervised learning, the parsing and production of natural language, pattern recognition, computer vision, information retrieval, several kinds of reasoning, planning, problem solving, and information compression.

\subsection{Foundations and Scope of the SP Theory}\label{sp_foundations_and_scope_appendix}

The SP theory is founded on a range of observations suggesting the fundamental importance of information compression in natural and artificial intelligence, in computing, in mathematics, and in neuroscience (\cite{sp_foundations}, \cite[Chapter 2]{wolff_2006}).

Like most theories, the SP theory is narrower in its scope than one might wish. It is certainly not a comprehensive theory of human psychology. For example, it has little to say about emotions and motivations and their impact, in people, on such things as perception, learning, and reasoning (but see \cite[Section V-A.2]{sp_autonomous_robots}). At some stage, there is likely to be a case for examining whether or how those kinds of things may be accommodated in the theory.

\subsection{The SP Computer Model and the SP Machine}\label{sp_computer_model_and_machine_appendix}

The SP theory is realised in the form of a computer model, SP71, which may be regarded as a version of the {\em SP machine}.

An outline of the organisation and workings of the SP computer model works may be found in \cite[Section 3.9]{wolff_2006}, with more detail, including pseudocode, in \cite[Sections 3.10 and 9.2]{wolff_2006}.\footnote{These sources describe SP70, a slightly earlier version of the model than SP71 but quite similar to it. The description of SP70 includes a description, in \cite[Sections 3.9.1 and 3.10]{wolff_2006}, of a subset of the SP70 model called SP61.} Fully commented source code for the SP71 computer model may be downloaded via a link near the bottom of \href{http://www.cognitionresearch.org/sp.htm}{www.cognitionresearch.org/sp.htm}, and via `Ancillary files' under \href{http://arxiv.org/abs/1306.3888}{www.arxiv.org/abs/1306.3888}.

All the multiple alignments shown in this paper are output from the SP computer model.

\subsection{Patterns and Symbols}\label{patterns_and_symbols_appendix}

In the SP system, knowledge is represented with arrays of atomic symbols in one or two dimensions called {\em patterns}. The SP71 model works with one-dimensional patterns but it is envisaged that the system will be generalised to work with patterns in two dimensions \cite[Section 3.3]{sp_extended_overview}.

Each SP pattern has an associated frequency of occurrence that has a role in the calculation of probabilities, as outlined in Appendix \ref{ic_prediction_probabilities_appendix}.

An `atomic symbol' in the SP system is simply a mark that can be matched with any other symbol to determine whether it is the same or different---no other result is permitted.

Patterns in two dimensions are likely to have a role in the processing of images (\cite[Chapter 13]{wolff_2006}, \cite{sp_vision}) and also in the processing of sensory or motor streams of information that occur in parallel \cite[Sections IV-A.4, IV-H, V-G to V-I, and Appendix C]{sp_autonomous_robots}.

In themselves, SP patterns are not particularly expressive. But within the multiple alignment framework (Appendix \ref{ma_appendix}), they support the representation and processing of a wide variety of kinds of knowledge (Section \ref{representation_processing_knowledge_section}, Appendix \ref{empirical_conceptual_support_appendix}). It appears that the system has potential as a {\em universal framework for the representation and processing of knowledge} (UFK) \cite[Section III]{sp_big_data}.

\subsection{Information Compression}\label{information_compression_appendix}

In the SP system, all kinds of processing is done by compression of information. This is essentially the principle of {\em minimum length encoding} (MLE) \cite{solomonoff_1964,wallace_boulton_1968,rissanen_1978}\footnote{MLE is an umbrella term for `minimum message length' encoding (MML), `minimum description length' encoding (MDL), and similar concepts.} but with qualifications described in Section \ref{mle_section}.

The default assumption in the SP theory is that compression of information is always lossless, meaning that all non-redundant information is retained. In particular applications, there may be a case for discarding non-redundant information (see, for example, \cite[Section X-B]{sp_big_data}) but any such discard is reversible.

The name `SP' is short for {\em Simplicity} and {\em Power}, because compression of any given body of information, {\bf I}, may be seen as a process of reducing `redundancy' of information in {\bf I} and thus increasing its `simplicity', whilst retaining as much as possible of its non-redundant descriptive and explanatory `power'. As noted in Appendix \ref{occams_razor_appendix}, it is no accident that the same two concepts are prominent in Occam's Razor as a touchstone of success for scientific theories.

In the SP system, information compression is achieved via the matching and unification of patterns, or parts thereof (see `ICMUP' in Section \ref{ovv_icmup_section} and Appendix \ref{icmup_appendix}). More specifically, it is achieved via the building of multiple alignments and via the unsupervised learning of grammars. These three things are described briefly in the following three subsections.

\subsubsection{Information Compression Via the Matching and Unification of Patterns}\label{icmup_appendix}

The basis for information compression in the SP system is a process of searching for patterns that match each other with a process of merging or `unifying' patterns that are the same: `information compression via the matching and unification of patterns' or `ICMUP' \cite{sp_foundations}.

At the heart of the SP computer model is a method for finding good full and partial matches between sequences, with advantages compared with classical methods \cite[Appendix A]{wolff_2006}.\footnote{The main advantages are \cite[Section 3.10.3.1]{wolff_2006}: 1) That it can match arbitrarily long sequences without excessive demands on memory; 2) For any two sequences, it can find a set of alternative matches (each with a measure of how good it is) instead of a single `best' match; 3) The `depth' or thoroughness of the searching, which has the effect of controlling the amount of backtracking, can be controlled by parameters.}

The emphasis on ICMUP is motivated partly by evidence of the importance of such processes in human perception and cognition, and partly by its potential to cut through much complexity and to achieve a new perspective on AI, mainstream computing, and mathematics \cite{sp_foundations}. Because a goal of the SP theory is to develop a new perspective on those four areas, without theoretical `baggage', the theory minimises the use of concepts from those disciplines, including mathematics \cite[Section 2.1]{sp_foundations}.\footnote{And, bearing in mind that the SP theory should be consistent with the biological origins of human intelligence, an attempt has been made to ensure that the frequency information that is stored with each SP pattern, and the probability calculations that are performed by the SP computer model, are, potentially, the kinds of things that could be modelled, at least approximately, via analogue processes in biological systems.}

This is not intended to be in any way disrespectful of mathematics as a discipline or the monumental achievements of mathematicians. It is merely an observation that, if one is trying to achieve a new perspective on any discipline, it is probably best to avoid using too many concepts from that discipline. With regard to mathematics, it would be surprising if it had not been shaped, at least in part, by its exceptionally long history, and the fact that, for most of its existence, the discipline has depended largely on human brainpower, with nothing but the simplest kinds of artificial aid.

\subsubsection{Information Compression Via the Building of Multiple Alignments}\label{ma_appendix}

The process for finding good full and partial matches between patterns is the foundation for processes that build {\em multiple alignments} like the one shown in Figure \ref{parsing_figure}. This concept is similar to multiple alignment in bioinformatics but with important differences \cite[Section 3.4]{wolff_2006}. It is a powerful and distinctive feature of the SP system.

\begin{figure*}[!htbp]
\fontsize{10.00pt}{12.00pt}
\centering
{\bf
\begin{BVerbatim}
0        1        2     3    4    5     6     7       8     9      10     11

                                        S
                                        SNG
                  NP ------------------ NP
                  Np1
                  D --- D
                        d2
a --------------------- a
                  #D -- #D
         N ------ N
         Ns --------------------------- Ns
         n6
stitch - stitch
         #N ----- #N
                  #NP ----------------- #NP
                                  Q --- Q
                                  q1
                             P -- P
                             p4
in ------------------------- in
                             #P - #P
                                  NP ------------------------------------ NP
                                                                          Np2
                                                                   N ---- N
                                                                   Ns
                                                                   n7
time ------------------------------------------------------------- time
                                                                   #N --- #N
                                  #NP ----------------------------------- #NP
                                  #Q -- #Q
                                        VP ---------- VP
                                                      Vp1
                                              V ----- V
                                        Vs -- Vs
                                              v2
saves --------------------------------------- saves
                                              #V ---- #V
                                                      N --- N
                                                            Ns
                                                            n8
nine ------------------------------------------------------ nine
                                                      #N -- #N
                                        #VP --------- #VP
                                        #S

0        1        2     3    4    5     6     7       8     9      10     11
\end{BVerbatim}
}
\caption{A multiple alignment illustrating parsing of a sentence, as described in the text.}
\label{parsing_figure}
\end{figure*}

% \newpage
% \addtolength{\topmargin}{.875in}

In Figure \ref{parsing_figure}, the SP pattern in column 0 is is a New pattern representing a sentence to be parsed, while each of columns 1 to 11 contains an Old SP pattern representing a grammatical form (where `grammatical form' includes words). This example is the best multiple alignment created by the SP computer model with the New pattern as shown in column 0 and a set of pre-existing Old patterns, including those shown in columns 1 to 11 in the figure.

Here, the `best' multiple alignment is the one in which the New pattern may be encoded most economically in terms of the Old patterns---and this means a multiple alignment in which there is a relatively large number of symbols that match each other from column to column, aligned in rows.

The way in which an encoding is derived from a multiple alignment is explained in \cite[Section 3.5]{wolff_2006} and \cite[Section 4.1]{sp_extended_overview}. Like all other kinds of knowledge in the SP system, encodings derived from multiple alignments are recorded using SP patterns (Appendix \ref{patterns_and_symbols_appendix}).

The overall effect of this multiple alignment is to analyse the sentence into its grammatical parts and sub-parts, an analysis that is, in its essentials, the same as a conventional parsing.

A point of interest is that the pattern for the whole sentence in column 6 marks the grammatical dependency between the singular subject of the sentence (`\texttt{a stitch}')---marked with `\texttt{Ns}'---and the singular verb (`\texttt{saves}')---marked with `\texttt{Vs}'. Notice how the dependency neatly bridges the subordinate phrase (`\texttt{in time}'). This method of encoding discontinuous dependencies in syntax contributes to the compression that is achieved by this multiple alignment and is, arguably, simpler than existing techniques for encoding such dependencies.

\subsubsection{Information Compression Via Unsupervised Learning}\label{ic_unsupervised_learning_appendix}

From a set of New patterns, the SP system may, without assistance from a `teacher' or anything equivalent, derive one or more plausible context-sensitive grammars, including segmental structures, classes of structure, and abstract patterns. The learning process is outlined in \cite[Section 5.1]{sp_extended_overview} and \cite[Section 3.9.2]{wolff_2006}, and described more fully in \cite[Chapter 9]{wolff_2006}. In that process, multiple alignment has a central role as a source of SP patterns for possible inclusion in any grammar (\cite[Section 9,2,5]{wolff_2006}, \cite[Section 5.1.1]{sp_extended_overview}). Although the current model has some shortcomings (Appendix \ref{future_developments_appendix}, \cite[Section 3.3]{sp_extended_overview}), it appears that these may be overcome.

A key part of the learning process is the formation of multiple alignments in which there are mismatches between patterns like those shown in Figure \ref{learning_alignment_figure}.

\begin{figure}[!htbp]
\fontsize{10.00pt}{12.00pt}
\centering
{\bf
\begin{BVerbatim}
0     t h a t g i r l r u n s    0
      | | | |         | | | |
1 A 1 t h a t b o y   r u n s #A 1
\end{BVerbatim}
}
\caption{A simple multiple alignment from which other patterns may be derived.}
\label{learning_alignment_figure}
\end{figure}

From the matched and unmatched parts of a multiple alignment like this, the system derives such patterns as `\texttt{B 1 t h a t \#B}', `\texttt{C 1 r u n s \#C}', `\texttt{D 1 g i r l \#D}', and `\texttt{D 2 b o y \#D}', each one with system-generated ID-symbols at the beginning and end. The system also creates a pattern like this: `\texttt{E 1 B \#B D \#D C \#C \#E}', that records the sequence of structures in terms of the ID-symbols. These patterns are the beginnings of a simple grammar.

In practice, many of the multiple alignments, and many of the derived segments, are much less tidy than this example may suggest. But, via heuristic search (Appendix \ref{heuristic_search_appendix}) through the space of alternative grammars, the system is able to derive one or two grammars that score relatively well in terms of information compression. Normally, these are also grammars that appear to be most natural, in accordance with the DONSVIC principle (Section \ref{ovv_donsvic_section}).

\subsubsection{Heuristic Search}\label{heuristic_search_appendix}

Like most problems in artificial intelligence, each of the afore-mentioned problems---finding good full and partial matches between patterns, finding or constructing good multiple alignments, and inferring one or more good grammars from a body of data---is normally too complex to be solved by exhaustive search.

With intractable problems like these, it is often assumed that the goal is to find theoretically ideal solutions. But with these and most other AI problems, ``The best is the enemy of the good''. By scaling back one's ambitions and searching for `reasonably good' solutions, it is often possible to find solutions that are useful, and without undue computational demands.

As with other AI applications, and as with the building of multiple alignments in bioinformatics, the SP71 model uses heuristic techniques---`hill climbing' or `descent'---in all three cases mentioned above \cite[Appendix A; Sections 3.9 and 3.10; Chapter 9]{wolff_2006}. This means searching for solutions in stages, with a pruning of the search tree at every stage, guided by measures of compression, and with backtracking where appropriate to increase the chances of success. With these kinds of techniques, acceptably good approximate solutions can normally be found without excessive computational demands and with `big O' values that are within acceptable limits. Possible alternatives to the heuristic techniques used in SP71 include `simulated annealing' and `genetic programming'.

\subsubsection{Grammars and Encodings, Simplicity and Power}\label{grammar_encoding_simplicity_power_appendix}

In unsupervised learning in the SP system, compression of a body of information, {\bf I}, produces two distinct results: a {\em grammar} and an {\em encoding} of {\bf I} in terms of the grammar, both of them expressed as SP patterns. The two together represent a lossless compression of {\bf I}.

The term `grammar' has been adopted because the SP programme of research derives largely from earlier research on models of language learning and grammatical inference \cite{wolff_1988} but, because of the versatility of SP patterns in the multiple alignment framework (Appendix \ref{patterns_and_symbols_appendix}), the term is applied, in this research, to any kind of knowledge, not just natural language.

Often but not invariably, there is a trade-off between the size of the grammar and the size of the encoding: as a general rule, small grammars yield large encodings and large grammars yield small encodings. Normally, the greatest overall compression of {\bf I} is obtained with grammars that are not at the extremes of size (small or big), and likewise for encodings. It appears that this means learning that avoids both under-generalisation and over-generalisation (Sections \ref{dlnn_under_over_generalisation_section} and \ref{nl_learning_section}).

From the trade-off we can see that there is a direct relationship between the concepts of `grammar' and `encoding' on the one hand, and the concepts of `simplicity' and `power' on the other: for a given {\bf I}, there is simplicity in any grammar when the grammar is small, and the grammar has power when the encoding is small. Any reasonably thorough compression of {\bf I} is likely to yield a good balance between the two.\footnote{Here, the qualification, ``reasonably thorough'' is quite important. Compression algorithms like the popular LZ algorithms are `quick and dirty'---they are designed for speed on low-powered computers---they are not very thorough and will normally miss quite large amounts of redundancy.}

\subsection{Information Compression, Prediction, and Probabilities}\label{ic_prediction_probabilities_appendix}

Owing to the close connection between information compression and concepts of prediction and probability \cite{li_vitanyi_2014}, the SP system is fundamentally probabilistic. As noted in Appendix \ref{patterns_and_symbols_appendix}, each SP pattern has an associated frequency of occurrence. Using that frequency information, robabilities may be calculated for each multiple alignment and for any inference that may be drawn from any given multiple alignment \cite[Section 3.7]{wolff_2006}.

Although the SP system is fundamentally probabilistic: it can be constrained to answer only those kinds of questions where probabilities are close to 0 or 1; and, via the use of error-reducing redundancy, it can deliver decisions with high levels of confidence. Contrary to what one may suppose, there is no conflict between the use of error-reducing redundancy and the notion that `computing' may be understood as information compression. The two things are independent, as described in \cite[Section 2.3.7]{wolff_2006}.

\subsection{SP-Neural}\label{sp-neural_appendix}

Part of the SP theory is the idea, described most fully in \cite[Chapter 11]{wolff_2006}, that the abstract concepts of {\em symbol} and {\em pattern} in the SP theory may be realised more concretely in the brain with collections of neurons in the cerebral cortex.\footnote{See also \cite[Section 2.3.1]{wolff_2006}.}

The neural equivalent of an SP pattern is called a {\em pattern assembly}. The word `assembly' has been adopted in this term because the concept is quite similar to Hebb's \cite{hebb_1949} concept of a {\em cell assembly}. The main difference is that the concept of pattern assembly is unambiguously explicit in proposing that the sharing of structure between two or more pattern assemblies is achieved by means of `references' from one structure to another, as described and discussed in \cite[Section 11.4.1]{wolff_2006}). Also, learning in the SP system is quite different from the gradualist kinds of learning that are popular in artificial neural networks (see Sections \ref{dlnn_scope_for_adaptation_section} and  \ref{dlnn_one-trial_learning_section}).

\begin{figure*}[!htbp]
\centering
\includegraphics[width=0.8\textwidth]{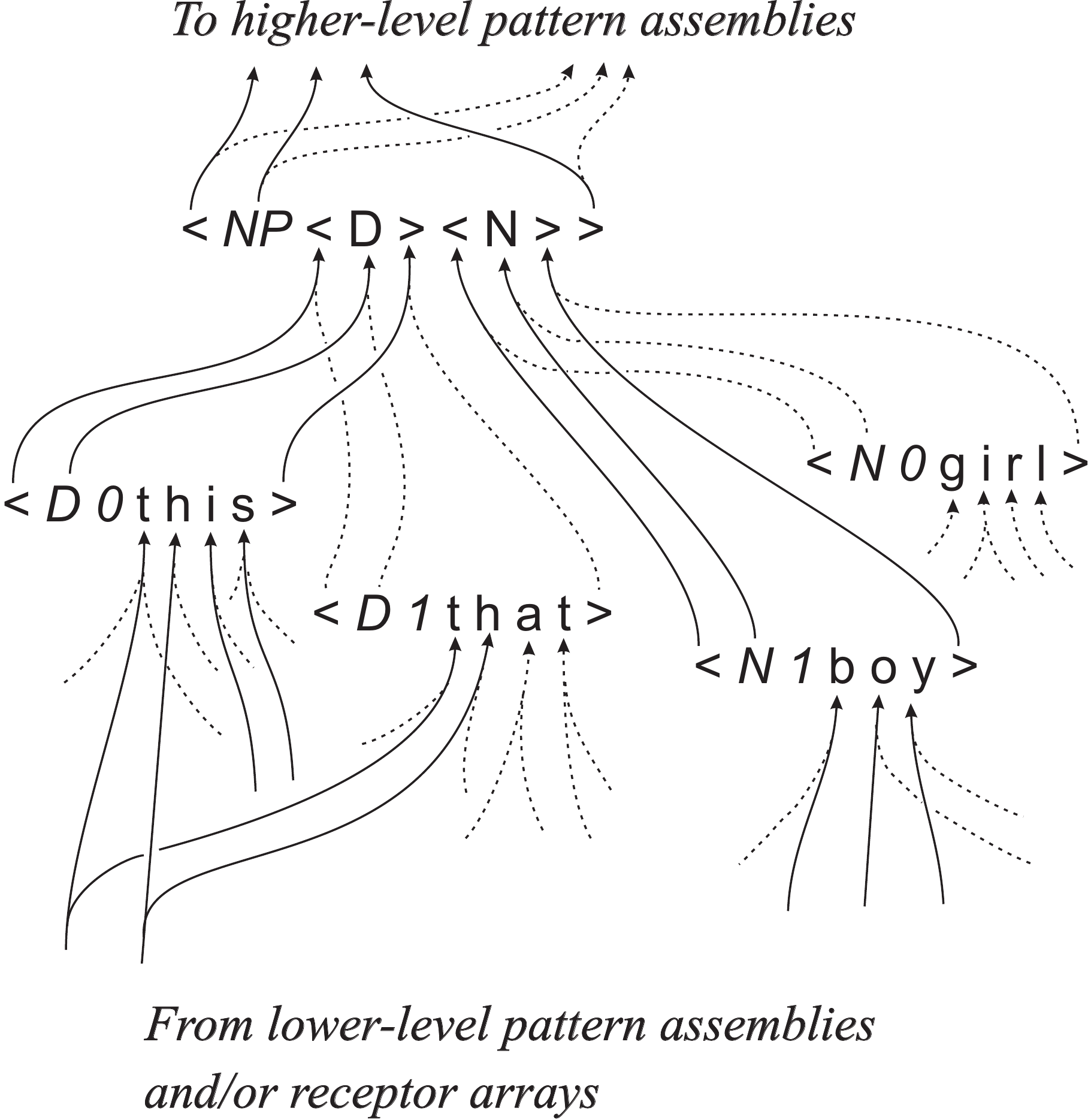}
\caption{Schematic representation of inter-connections amongst pattern assemblies as described in the text. Not shown in the figure are lateral connections within each pattern assembly, and inhibitory connections elsewhere, as outlined in \cite[Sections 11.3.3 and 11.3.4]{wolff_2006}. Reproduced, with permission, from Figure 11.2 in \cite{wolff_2006}.}
\label{connections_figure}
\end{figure*}

Figure \ref{connections_figure} shows schematically how pattern assemblies may be represented and inter-connected with neurons. Here, each pattern assembly, such as `\texttt{< NP < D > < N > >}', is represented by the sequence of atomic symbols of the corresponding SP pattern. Each atomic symbol, such as `\texttt{<}' or `\texttt{NP}', would be represented in the pattern assembly by one neuron or a small group of inter-connected neurons. Apart from the inter-connections amongst pattern assemblies, the cortex in SP-neural is somewhat like a sheet of paper on which knowledge may be written in the form of pattern assemblies and their inter-connections.

As noted in Section \ref{dlnn_biological_validity_section}, the hierarchical relations that can be seen in Figure \ref{connections_figure} may be seen to be broadly in keeping with the hierarchical relations between `simple' and `complex' cells discovered by Hubel and Wiesel \cite{hubel_2000}.

It is envisaged that any pattern assembly may be `recognised' if it receives more excitatory inputs than rival pattern assemblies, perhaps via a winner-takes-all mechanism in which inhibitory processes have a role to play \cite[Section 11.3.4]{wolff_2006}. And, once recognised, any pattern assembly may itself be a source of excitatory signals leading to the recognition of higher-level pattern assemblies.

\subsection{Empirical and Conceptual Support for the SP Theory}\label{empirical_conceptual_support_appendix}

As noted in Appendix \ref{sp_simplicity_power_evaluation_appendix}, the SP theory has non-trivial things to say about a wide range of observations and concepts in artificial intelligence, mainstream computing, mathematics, and human perception and cognition. These things are described most fully in \cite{wolff_2006}, more briefly in \cite{sp_extended_overview}, and in extended summaries in \cite[Sections IV and V]{sp_autonomous_robots}. In a bare-bones summary, the main strengths of the SP system are in:

\begin{itemize}

\item Natural language processing (\cite[Chapter 5]{wolff_2006}, \cite[Section 8]{sp_extended_overview}).

\item Pattern recognition and vision (\cite[Chapter 6]{wolff_2006}, \cite[Section 9]{sp_extended_overview}, \cite{sp_vision}).

\item Information storage and retrieval (\cite[Chapter 6]{wolff_2006}, \cite[Section 11]{sp_extended_overview}, \cite{wolff_sp_intelligent_database}).

\item The representation and processing of diverse kinds of knowledge (\cite[Section 7]{sp_extended_overview}, \cite[Section III-B]{sp_big_data} and, more generally, \cite[Chapters 5 to 10]{wolff_2006}).

\item Benefits accruing from the seamless integration of diverse kinds of knowledge and diverse aspects of intelligence (\cite[Sections 2, 5, and 7]{sp_benefits_apps}).

\item Several kinds of reasoning (\cite[Chapter 7]{wolff_2006}, \cite[Section 10]{sp_extended_overview}).

\item Planning and problem solving (\cite[Chapter 8]{wolff_2006}, \cite[Section 12]{sp_extended_overview}).

\item Unsupervised learning (\cite[Chapter 9]{wolff_2006}, \cite[Section 5]{sp_extended_overview}, \cite[Section V]{sp_autonomous_robots}).

\item Implications for our understanding of human perception and cognition, including neural processing (\cite[Chapters 11 and 12]{wolff_2006}, \cite{sp_vision}).

\item Implications for our understanding of the nature of mathematics (\cite[Chapter 10]{wolff_2006}, \cite{sp_foundations}).

\end{itemize}

There is more detail about some of these capabilities in the body of the paper.

\subsection{Potential Benefits and Applications}\label{benefits_applications_appendix}

In summary, potential benefits and applications of the SP system include:

\begin{itemize}

\item Helping to solve nine problems associated with big data \cite{sp_big_data} (see also Section \ref{big_data_autonomous_robots_section}).

\item The development of versatility and adaptability in autonomous robots, with potential for gains in computational efficiency \cite{sp_autonomous_robots} (see also Section \ref{big_data_autonomous_robots_section}).

\item The development of computer vision and pattern recognition, and the interpretation of aspects of natural vision (\cite{sp_vision}, \cite[Section 9]{sp_extended_overview}).

\item The system may be developed as a versatile database management system, with intelligence \cite{wolff_sp_intelligent_database}.

\item The system may serve as a repository for medical knowledge and as an aid for medical diagnosis \cite{wolff_medical_diagnosis}.

\item There are several other potential benefits and applications described in \cite{sp_benefits_apps}: simplification of computing systems, including software;  unsupervised learning; the processing of natural language; software engineering; information compression; the semantic web; bioinformatics; the detection of computer viruses; data fusion; new kinds of computer; the development of scientific theories; and the seamless integration of diverse kinds of knowledge and processing.

\end{itemize}

As describe in Appendix \ref{future_developments_appendix}, next, some potential applications may be developed on relatively short timescales.

\subsection{Development of the SP System}\label{future_developments_appendix}

Like most scientific theories, the SP system is not complete \cite[Section 3.3]{sp_extended_overview}. As it is now, the main shortcomings in the SP computer model are:

\begin{itemize}

\item The process for finding good full and partial matches between one-dimensional patterns needs to be generalised to patterns in two dimensions;

\item A better understanding is needed of how the system may be applied to the discovery and recognition of low-level features in speech and images;

\item In unsupervised learning, the model does not learn intermediate levels of abstraction or discontinuous dependencies in data;

\item And a better understanding is needed of how the system may be applied in the representation and processing of numbers.

\end{itemize}

\noindent It appears that none of these problems are fatal---that all of them are soluble.

Apart from this work, developing the core model, there is much to be done in exploring how the system may be applied in areas such as pattern recognition, natural language processing, and so on (Appendix \ref{empirical_conceptual_support_appendix}), and in applications such as medical diagnosis, crime prevention and detection, and more (Appendix \ref{benefits_applications_appendix}).

Since there are many more avenues to be explored than could be tackled by any one research group, it is envisaged that the SP computer model will be the basis for the creation of a high-parallel, open-source version of the SP machine, hosted on an existing high-performance computer \cite{sp_proposal}. This will be a means for researchers everywhere to explore what can be done with the system, and to create new versions of it. If any reader is interested in the possibility of using such a facility or believes the facility may be useful, it would be helpful if he or she were to let the author know (via \href{mailto:jgw@cognitionresearch.org}{jgw@cognitionresearch.org}).

Some potential applications of the SP system may be developed on relatively short timescales using existing high-performance computers or even ordinary computers. These include the SP system as an intelligent database \cite{wolff_sp_intelligent_database}, and applications in such areas as medical diagnosis \cite{wolff_medical_diagnosis}, pattern recognition (\cite[Chapter 6]{wolff_2006}, \cite[Section 9]{sp_extended_overview}), information compression \cite[Section 6.7]{sp_benefits_apps}, highly-economical transmission of information \cite[Section VIII]{sp_big_data}, bioinformatics \cite[Section 6.10.2]{sp_benefits_apps}, and natural language processing \cite[Section 6.2]{sp_benefits_apps}.

\section{Occam's Razor: Simplicity and Power}\label{occams_razor_appendix}

One of the most widely accepted principles in science---Occam's Razor---is that a good theory should combine conceptual {\em simplicity} with explanatory or descriptive {\em power}. Albert Einstein expressed it thus: ``A theory is more impressive the greater the simplicity of its premises, the more different things it relates, and the more expanded its area of application.''\footnote{Quoted in \cite[p.~512]{isaacson_2007}}

In these terms, a theory can be weak because it is too complex---like the Ptolemaic earth-centred system of epicycles as a theory of the movements of the planets and the sun, or merely redescribing the data that it is meant to explain. Or a theory can be weak because it is too simple and too general, explaining everything and nothing---like the over-enthusiastic use of the concept of `instinct' as an explanation of animal behaviour. A good theory---like the Copernicus/Kepler heliocentric theory of the solar system---strikes a balance between the two. This relates to the issue of under-generalisation and over-generalisation discussed in Sections \ref{dlnn_under_over_generalisation_section} and \ref{nl_learning_section}.

Alan Turing's concept of a `Universal Turing Machine', and equivalent models such as Post's \cite{post_1943} `Canonical System', are good models of `computing' in the widest sense. But notwithstanding Turing's vision that computers might become intelligent \cite{turing_1950}, the concept of a Universal Turing Machine, does not tell us how!\footnote{But Turing began to address that problem in a report about ``unorganised machines'' \cite{turing_1992} and also in \cite{turing_1950}.} Something with more specific capabilities is needed for AI.

Since the beginnings of electronic computers, several decades of research in AI have yielded some useful insights and some impressive applications but I believe it is fair to say that AI has been and is suffering from an excess of narrow subfields and, with some honourable exceptions, insufficient attention to the need to simplify and integrate observations and concepts across different areas.\footnote{Honourable exceptions include research aiming to develop `unified theories of cognition' and `artificial general intelligence'.} Hence the SP programme of research (Appendix \ref{motivation_appendix}).

It is no accident that `simplicity' and `power'---key ideas in evaluating scientific theories---are also prominent in the SP theory:

\begin{itemize}

\item The two terms, together, are equivalent to `information compression', which is central in the SP theory (Appendix \ref{information_compression_appendix});

\item Cosmologist John Barrow has written that ``Science is, at root, just the search for compression in the world'' \cite[p.~247]{barrow_1992}.

\end{itemize}

\subsection{Motivation}\label{motivation_appendix}

Part of the motivation for developing the SP theory has been to try to overcome the problems of narrow focus and over-specialisation, mentioned above and identified by other authors:

\begin{itemize}

\item Neisser \cite{neisser_1967} writes of the need to avoid `microtheories' in psychology.

\item In a similar vein, Newell, in his famous essay ``You can't play 20 questions with nature and win'' \cite{newell_1973}, urges researchers to develop theories with wide scope (pp.~284--289) dealing with ``a genuine slab of human behaviour'' (p.~303). This thinking led on to Newell's {\em Unified Theories of Cognition} \cite{newell_1990} and related work, discussed in Section \ref{utc_agi_section}.

\item ``Today, as scientists labor to create machine technologies to augment our senses, there's a strong tendency to view each sensory field in isolation as specialists focus only on a single sensory capability. Experts in each sense don't read journals devoted to the others senses, and they don't attend one another's conferences. Even within IBM, our specialists in different sensing technologies don't interact much.'' \cite[location 1004]{kelly_hamm_2013}.

\item And McCorduck writes ``The goals once articulated with debonair intellectual verve by AI pioneers appeared unreachable ...~Subfields broke off---vision, robotics, natural language processing, machine learning, decision theory---to pursue singular goals in solitary splendor, without reference to other kinds of intelligent behaviour.'' \cite[p.~417]{mccorduck_2004}. Later, she writes of ``the rough shattering of AI into subfields ...~and these with their own sub-subfields---that would hardly have anything to say to each other for years to come.'' ({\em ibid.}, p.~424). She adds: ``Worse, for a variety of reasons, not all of them scientific, each subfield soon began settling for smaller, more modest, and measurable advances, while the grand vision held by AI's founding fathers, a general machine intelligence, seemed to contract into a negligible, probably impossible dream.''

\end{itemize}

\subsection{Evaluation of the Sp Theory in Terms of Simplicity and Power}\label{sp_simplicity_power_evaluation_appendix}

Although, in comparing one theory with another, we must rely on relatively informal assessments of simplicity and power, the SP theory, in those terms, appears to do well \cite[Section 4]{sp_benefits_apps}:

\begin{itemize}

\item The key concept of multiple alignment, with associated processes, are, in the SP computer model, expressed in an `exec' file that requires less than 500 KB of storage space.

\item The theory has non-trivial things to say about a wide range of observations and concepts in artificial intelligence, mainstream computing, mathematics, and human perception and cognition (Appendix \ref{empirical_conceptual_support_appendix}) and it has many potential benefits and applications (Appendix \ref{benefits_applications_appendix}).

\end{itemize}

It appears that the SP theory avoids what are perhaps the two most common pitfalls in the development of scientific theories, noted above: it is not over-specific and it is not over-general. There appears to be a good balance between simplicity and power.

\subsection{If It Works, Who Cares?}\label{if_it_works_appendix}

Some people may argue that these concerns are misplaced---that researchers should concentrate on creating things that work and not worry about the development of good theory. It is true that a suck-it-and-see approach can produce useful results and may indeed be helpful in the development of theory. But no theory or bad theory is almost always a handicap: imagine the difficulties of space travel using Ptolemy's epicycles as a guide, or the impoverishment of biology without an understanding of DNA.

% There is more about these issues in Section \ref{watson_section}.

\subsection{Since the Human Mind Is a Kluge, Why Worry About Good Theory for AI?}\label{kluge_appendix}

Marcus has argued persuasively \cite{marcus_2008} that, as a result of the way biological evolution builds on what comes to hand, the human mind is, in many respects, a kluge, without the coherence or elegance of a well-designed piece of engineering. In keeping with that idea, Minsky \cite{minsky_1986} writes: ``What magical trick makes us intelligent? The trick is that there is no trick. The power of intelligence stems from our vast diversity, not from any single, perfect principle.'' (p.~308).

Since human perception and cognition is a source of inspiration and a touchstone of success for AI, some people may conclude from arguments like those just mentioned that we need not worry about good theory for AI: just bundle together a collection of applications from different areas of AI. But this is really a counsel of despair which should not distract us:

\begin{itemize}

\item Although it is clear that the human mind has many shortcomings, it also has extraordinary versatility and adaptability. This is still a major challenge for AI, a challenge which appears to demand the development of good theory.

\item A kluge of different applications is unlikely to provide the smooth inter-working of different kinds of knowledge and different aspects of intelligence that appears to be essential if, in AI, we are to achieve human-like versatility and adaptability in intelligence (Section \ref{ovv_simplification_integration_computers_section}).

\item Much progress in science has depended on a willingness to look for simplicity within the apparent complexity of the world, witness Newton's laws of motion. We should not give up on the quest for good theory merely because it is difficult.

\end{itemize}

Of course, it is known that different areas of the brain are specialised for different functions. But there is no contradiction between that observation and the likelihood that human intelligence depends on powerful general principles. The SP concepts are promising candidates for those roles.

\section{Long-Term and Short-Term Perspectives}\label{long_short_perspectives_appendix}

In research and development in ICT, it is often possible to achieve useful results on relatively short timescales. For example, Google Translate delivers translations between natural languages that are often useful even though they normally fall short of what can be achieved by a good human translator.

To achieve higher standards and crack harder problems, an over-enthusiastic focus on short-term results can be more costly than a more strategic approach. Any one idea that shows a little promise may turn out to be a blind alley (in the same way that climbing a tree, while it takes us closer to the moon, is not a means of reaching the moon \cite{dreyfus_1992}). Big bets on ideas like that can mean big losses. It is probably better to maintain a broad view unless or until there is sufficient evidence to justify a narrower focus.

The SP system, itself the product of a 20-year programme of research, draws on earlier research developing computer models of language learning. There is now much evidence in support of the framework but, as noted in Appendix \ref{future_developments_appendix}, there are still many avenues to be explored. Hence the proposal to create a new research facility to enable researchers anywhere to explore what can be done with the SP machine and to create new versions of it. Such a development would be a relatively inexpensive way of taking things forward on a broad front, very much in keeping with the quote from \cite{kelly_hamm_2013} in Section \ref{watson_as_kluge_section}: ``The creation of this new era of [cognitive] computing is a monumental endeavor ...''.

\bibliographystyle{plain}
% \bibliography{latex_references}

\end{document}